\documentclass[dvipsnames]{article}
\usepackage{frontiermat_arxiv}

\usepackage{iftex}
\ifPDFTeX
  \usepackage[utf8]{inputenc}
  \usepackage[T1]{fontenc}
\else
  \usepackage{fontspec}
\fi
\usepackage{amsmath,amssymb,amsfonts}
\usepackage{booktabs}
\usepackage{array}
\usepackage{multirow}
\usepackage{makecell}
\usepackage{tabularx}
\usepackage{colortbl}
\usepackage{subcaption}
\usepackage{placeins}
\usepackage{float}
\usepackage[all]{hypcap}
\usepackage{setspace}
\usepackage{ragged2e}
\usepackage{enumitem}
\usepackage{pifont}
\usepackage{xspace}
\usetikzlibrary{positioning,arrows.meta,fit,calc}
\usepackage[capitalise,noabbrev]{cleveref}


\newcommand{\bench}{FrontierMat\xspace}

\newcommand{\benchlogo}{\raisebox{-0.18\height}{\includegraphics[height=1.1em]{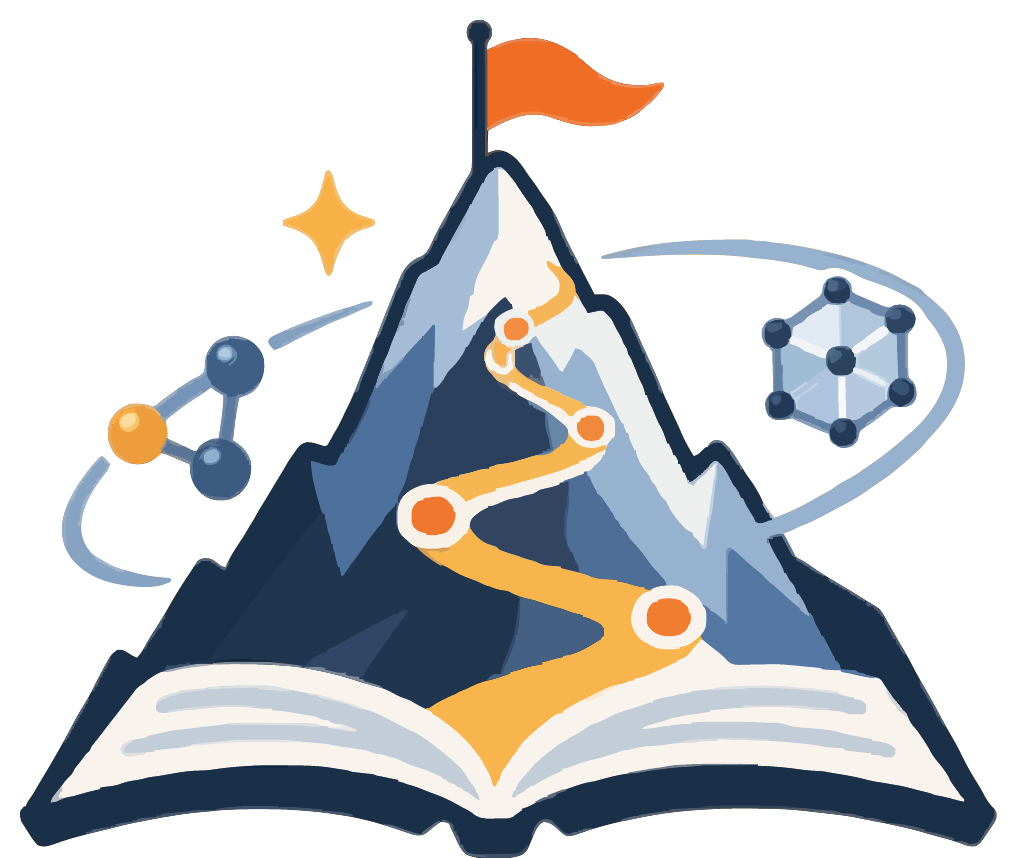}}}

\newcommand{\cmark}{\ding{51}}
\newcommand{\xmark}{\ding{55}}

\xspaceaddexceptions{\%]}

\newif\ifshowtodos
\showtodosfalse

\ifshowtodos
  \newcommand{\todo}[1]{{\color{red!75!black}\textbf{[TODO:} #1\textbf{]}}}
  
  \newcommand{\owner}[2]{{\color{blue!70!black}\footnotesize[#1 $\cdot$ #2]}}
  \newcommand{\note}[1]{{\color{teal!70!black}\footnotesize[Note: #1]}}
\else
  \newcommand{\todo}[1]{}
  
  \newcommand{\owner}[2]{}
  \newcommand{\note}[1]{}
\fi

\newenvironment{examplebox}[1]{%
  \par\smallskip\noindent\begin{tcolorbox}[colback=black!2,colframe=black!45,boxrule=0.4pt,
    arc=1.5pt,left=4pt,right=4pt,top=3pt,bottom=3pt,breakable,
    title={\footnotesize\bfseries #1},fonttitle=\footnotesize,colbacktitle=black!12,coltitle=black]\raggedright}%
  {\end{tcolorbox}}

\newif\ifshowowners
\showownersfalse
\definecolor{ownYou}{gray}{0.35}
\definecolor{ownA}{RGB}{31,94,170}
\definecolor{ownB}{RGB}{26,127,72}
\newcommand{\ownedby}[2]{%
  \ifshowowners\ifshowtodos\par\noindent{\scriptsize\sffamily
    \ifx#1A\colorbox{ownA!12}{\textcolor{ownA}{\textbf{Intern A}~\textbar~\texttt{#2}}}\else
    \ifx#1B\colorbox{ownB!12}{\textcolor{ownB}{\textbf{Intern B}~\textbar~\texttt{#2}}}\else
    \colorbox{ownYou!10}{\textcolor{ownYou}{\textbf{Lead author}~\textbar~\texttt{#2}}}\fi\fi}\par\fi\fi}

\definecolor{findframe}{HTML}{217D82}
\definecolor{findback}{HTML}{EEF5F5}
\newcounter{finding}
\newenvironment{finding}{%
  \refstepcounter{finding}\par\smallskip\noindent
  \begin{tcolorbox}[enhanced,colback=findback,colframe=findframe,boxrule=0.5pt,arc=2pt,
    left=4pt,right=4pt,top=2pt,bottom=2pt,before skip=3pt,after skip=4pt]\small
  \textbf{\textcolor{findframe}{Finding~\thefinding:}}~\itshape\ignorespaces}%
  {\end{tcolorbox}}

\newcommand{\NumGroups}{13\xspace}

\newcommand{\NumAcademicGroups}{9\xspace}
\newcommand{\NumIndustryGroups}{4\xspace}

\newcommand{\NumAll}{330\xspace}

\newcommand{\AllYearMin}{1998\xspace}
\newcommand{\AllYearMax}{2026\xspace}

\newcommand{\NumLitCand}{25{,}182\xspace}
\newcommand{\NumLitScreened}{4{,}000\xspace}
\newcommand{\NumLitPool}{2{,}186\xspace}

\newcommand{\NumAllSingleAns}{98\xspace}
\newcommand{\NumAllMultiAns}{232\xspace}

\newcommand{\NumAllMicro}{88\xspace}
\newcommand{\NumAllMeso}{24\xspace}
\newcommand{\NumAllMacro}{88\xspace}
\newcommand{\NumAllSingle}{200\xspace}
\newcommand{\NumAllCross}{130\xspace}

\newcommand{\OursModel}{\bench-Model\xspace}
\newcommand{\OursAcc}{55.4\xspace}
\newcommand{\BaseAcc}{36.7\xspace}
\newcommand{\AdaptGain}{18.7\xspace}
\newcommand{\BaseTrunc}{7.6\xspace}

\newcommand{\BaseFactAcc}{16.1\xspace}
\newcommand{\OursFactAcc}{38.3\xspace}
\newcommand{\BestModel}{GPT-6-Astra\xspace}
\newcommand{\BestAcc}{53.7\xspace}
\newcommand{\BestCILow}{48.9\xspace}
\newcommand{\BestCIHigh}{58.4\xspace}
\newcommand{\SolAcc}{51.4\xspace}

\newcommand{\OpusAcc}{47.3\xspace}

\newcommand{\DSAcc}{47.6\xspace}

\newcommand{\GLMAcc}{45.0\xspace}

\newcommand{\NumScaleComparison}{265\xspace}
\newcommand{\GenAvgBase}{79.0\xspace}
\newcommand{\GenAvgOurs}{79.8\xspace}
\newcommand{\MMLUDrop}{2.3\xspace}

\newcommand{\CrossGap}{8.7\xspace}
\newcommand{\CrossGapP}{0.008\xspace}
\newcommand{\CrossGapMin}{4.6\xspace}
\newcommand{\CrossGapMax}{11.3\xspace}

\newcommand{\DistElec}{39\xspace}

\newcommand{\DistMechShare}{40\xspace}

\newcommand{\DistElecCross}{24\xspace}

\newcommand{\MaScQABase}{94.8\xspace}
\newcommand{\MaScQAOurs}{94.9\xspace}

\newcommand{\AblBaseHO}{35.9\xspace}  \newcommand{\AblBaseFM}{18.0\xspace}
\newcommand{\AblMetaHO}{39.4\xspace}  \newcommand{\AblMetaFM}{18.3\xspace}
\newcommand{\AblMixHO}{59.0\xspace}   \newcommand{\AblMixFM}{20.1\xspace}
\newcommand{\AblSoEHO}{59.4\xspace}   \newcommand{\AblSoEFM}{20.5\xspace}
\newcommand{\AblMetaGainHO}{3.5\xspace}   \newcommand{\AblMetaCIHO}{[1.1, 5.9]\xspace}
\newcommand{\AblMetaGainFM}{0.3\xspace}   \newcommand{\AblMetaCIFM}{[$-$0.6, 1.1]\xspace}
\newcommand{\AblSoEGainHO}{23.5\xspace}   \newcommand{\AblSoECIHO}{[20.1, 26.9]\xspace}
\newcommand{\AblSoEGainFM}{2.5\xspace}    \newcommand{\AblSoECIFM}{[0.5, 4.6]\xspace}
\newcommand{\AblVsMixHO}{0.4\xspace}      \newcommand{\AblVsMixCIHO}{[$-$2.0, 2.9]\xspace}
\newcommand{\AblVsMixFM}{0.4\xspace}      \newcommand{\AblVsMixCIFM}{[$-$1.2, 2.0]\xspace}
\newcommand{\AblLossBase}{1.8549\xspace}        \newcommand{\AblLossMeta}{1.8529\xspace}
\newcommand{\NumAblHeldOut}{513\xspace}  \newcommand{\NumAblProbe}{786\xspace}  \newcommand{\NumAblBuckets}{99\xspace}

\newcommand{\MSQABase}{65.2\xspace}  \newcommand{\MSQAOurs}{73.8\xspace}  \newcommand{\MSQAGain}{8.6\xspace}
\newcommand{\MSBBase}{58.4\xspace}   \newcommand{\MSBOurs}{72.4\xspace}   \newcommand{\MatSciBenchGain}{14.0\xspace}
\newcommand{\MSQABest}{72.9\xspace}  \newcommand{\MSBBest}{75.2\xspace}  \newcommand{\MaScQABest}{62.0\xspace}
\newcommand{\MSBGapBest}{2.8\xspace}

\newcommand{\NumMultiAns}{232\xspace}
\newcommand{\GoldK}{2.33\xspace}
\newcommand{\OverOurs}{5\xspace}

\newcommand{\OverAstra}{19\xspace}
\newcommand{\UnderAstra}{30\xspace}

\newcommand{\KOpus}{3.03\xspace}

\newcommand{\OverShareErr}{49\xspace}
\newcommand{\AddWrongShareErr}{83\xspace}
\newcommand{\OverMinOthers}{35\xspace}
\newcommand{\OverMaxOthers}{49\xspace}

\newcommand{\NumSolvedNone}{129\xspace}
\newcommand{\NumSolvedAll}{105\xspace}
\newcommand{\ShareSolvedNone}{39\xspace}

\newcommand{\NumSolvedAtMostOne}{164\xspace}
\newcommand{\ShareSolvedAtMostOne}{50\xspace}
\newcommand{\UnsolvedDomMin}{17\xspace}
\newcommand{\UnsolvedDomMax}{56\xspace}

\newcommand{\NumSolvedNoneConsensus}{81\xspace}

\newcommand{\NumStratum}{202\xspace}
\newcommand{\StratMicro}{29.5\xspace}

\newcommand{\StratMeso}{32.7\xspace}

\newcommand{\StratMacro}{23.0\xspace}

\newcommand{\StratCross}{17.5\xspace}

\newcommand{\StratCrossLowestCount}{4\xspace}
\newcommand{\StratElec}{19.4\xspace}

\newcommand{\TradeoffStrat}{19.8\xspace}

\newcommand{\BootTopAstra}{97\xspace}

\newcommand{\BootTier}{94\xspace}

\newcommand{\NumInformative}{177\xspace}
\newcommand{\ShareDiscPos}{73\xspace}
\newcommand{\NumConstant}{153\xspace}
\newcommand{\MedianDisc}{0.36\xspace}

\renewenvironment{examplebox}[1]{%
  \par\smallskip\noindent\begin{tcolorbox}[enhanced,colback=fmTeal!4,colframe=fmTeal,boxrule=0.5pt,
    arc=2pt,left=5pt,right=5pt,top=3pt,bottom=3pt,breakable,
    title={\footnotesize\bfseries #1},fonttitle=\footnotesize,colbacktitle=fmTeal!14,coltitle=fmNavy]\raggedright}%
  {\end{tcolorbox}}

\newlength{\colfigwidth}
\title{FrontierMat: Benchmarking Language Models\\[2pt] at the Frontier of Materials R\&D}
\author{%
  \normalsize
  Feng Chen$^{1,2*}$\quad Jianguo Huang$^{1,2*}$\quad Yushan Xiao$^{2*}$\quad Yue Li$^{1*}$\quad Bijun Tang$^{2*}$\\
  Xuyu Dong$^{4,1}$\quad Xingguo Wang$^{1}$\quad Feiyang Xie$^{5}$\quad Jianhui Gu$^{5}$\\
  Dongfeng Cao$^{6\dagger}$\quad Hongda Chen$^{6}$\quad Yongji Gong$^{7}$\quad Ruiqi Li$^{8}$\quad Shuxin Li$^{6}$\\
  Xingji Li$^{9}$\quad Ruitao Lv$^{10}$\quad Li Song$^{11}$\quad Lei Wang$^{15}$\quad Ye Zhang$^{12,13}$\\
  Xiaoxu Zhao$^{3}$\quad Xitao Zheng$^{14\dagger}$\quad Bo An$^{1}$\quad Zheng Liu$^{1}$\\[0.9em]
  {\footnotesize
  $^{1}$Nanyang Technological University, Singapore\quad
  $^{2}$Shanghai Mataria Technology Co., Ltd.\\
  $^{3}$Peking University\quad
  $^{4}$Shanghai Jiao Tong University\quad
  $^{5}$Xi'an Jiaotong University\\
  $^{6}$Wuhan University of Technology\quad
  $^{7}$Beihang University\\
  $^{8}$Eighth Research Institute of Nuclear Industry, CNNC\quad
  $^{9}$Harbin Institute of Technology\\
  $^{10}$Tsinghua University\quad
  $^{11}$University of Science and Technology of China\\
  $^{12}$Commercial Aircraft Corporation of China (COMAC)\quad
  $^{13}$Tongji University\\
  $^{14}$Northwestern Polytechnical University\quad
  $^{15}$Wanhua Chemical Group Co., Ltd.\\[0.5em]
  Correspondence: \texttt{feng012@e.ntu.edu.sg}, \texttt{xiaoyushan@mataria.ai}, \texttt{yueli@ntu.edu.sg},\\
  \texttt{bjtang@mataria.ai}, \texttt{boan@ntu.edu.sg}, \texttt{Z.Liu@ntu.edu.sg}}%
}
\newcommand{\authornotes}{%
  \begingroup
  \renewcommand{\thefootnote}{}%
  \footnotetext{$^{*}$Equal contribution.}%
  \footnotetext{$^{\dagger}$Authors from Dongfeng Cao through Xitao Zheng contributed as expert reviewers from academia and industry and are listed alphabetically by surname.}%
  \endgroup}

\begin{document}
\maketitle
\authornotes

\begin{abstract}
To accelerate materials discovery, language models must be able to address problems at the research frontier. Yet textbook- and examination-derived benchmarks primarily assess established knowledge, leaving performance on frontier research problems largely untested. We introduce \bench, a benchmark of \NumAll problems drawn from frontier academic and industrial research in fields including aerospace, semiconductors, embodied intelligence and nuclear fusion. Every problem is validated by leading experimentalists and industry experts in its field. Each problem presents a real study and asks the model to infer a withheld outcome or draw a conclusion from selected experimental evidence. Problems are annotated by functional domain, material system and the scale of the governing mechanism; \NumAllCross require cross-scale reasoning. The strongest general-purpose model scores only \BestAcc\%, and on matched problems, cross-scale problems score \CrossGap points below single-scale ones. Half of the problems are solved by at most one model, and models fail chiefly by endorsing what is plausible rather than what was observed. A domain-adaptation study with meta-weighted continued pretraining and scale-modular self-distillation raises an open-weight model from \BaseAcc\% to \OursAcc\%, matching the strongest proprietary model, with improvements on public materials benchmarks and no decline in aggregate performance on general benchmarks. \bench provides a testbed for measuring progress toward language models that support frontier materials R\&D.
\end{abstract}

\section{Introduction}
\label{sec:intro}

\begin{figure}[t]
  \centering
  \includegraphics[width=0.8\textwidth]{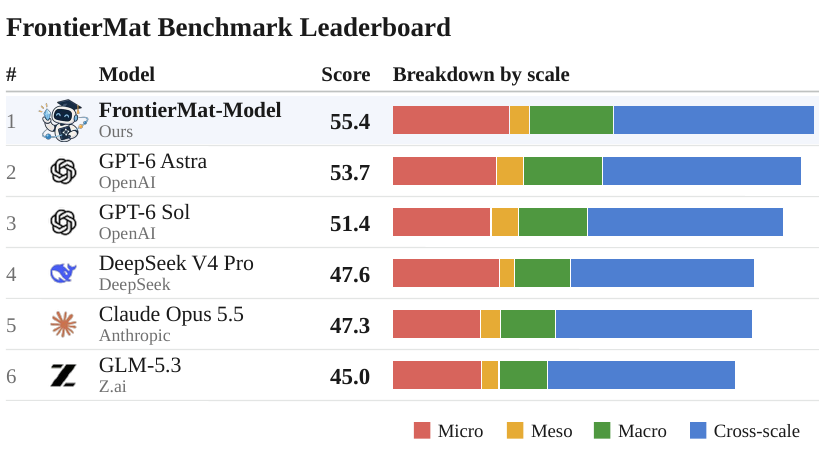}
  \caption{\bench leaderboard of six evaluated models. Bars split each overall score by the scale of the governing mechanism. Raw cross-scale scores are higher because these problems are more often single-answer; \S\ref{sec:analysis-scale} compares matched problems.}
  \label{fig:leaderboard}
\end{figure}

Lighter airframes, faster chips, dexterous robots and viable fusion reactors require materials advances pursued by academic laboratories and industrial R\&D teams. Language models are increasingly involved in this work: tool-augmented models already plan and run chemical experiments~\citep{boiko2023autonomous,bran2024chemcrow}, and their readiness for materials discovery is under active examination~\citep{miret2024ready}. Their value depends on addressing problems at this frontier. Such problems concern a specific material system under particular application constraints, often requiring researchers to connect defects and local structure to grains, fibres and pores, and ultimately to specimen or device performance. Answering them requires experimental evidence.

Current benchmarks provide limited evidence of this capability. Materials benchmarks draw on examinations and textbooks (MaScQA, \citealp{zaki2024mascqa}; MaterialBENCH, \citealp{yoshitake2024materialbench}; MatSciBench, \citealp{zhang2025matscibench}) or use language models to generate problems from papers, textbooks and databases (MSQA, \citealp{cheung2025msqa}; SciKnowEval, \citealp{feng2024sciknoweval}). Beyond materials, SuperGPQA~\citep{map2025supergpqa} screens candidate problems with model responses, and ChemBench~\citep{mirza2025chembench} labels difficulty by expert judgement. These benchmarks do not systematically combine problems from active R\&D with annotations of the mechanisms linking atomic structure to component performance, a central requirement of materials design.

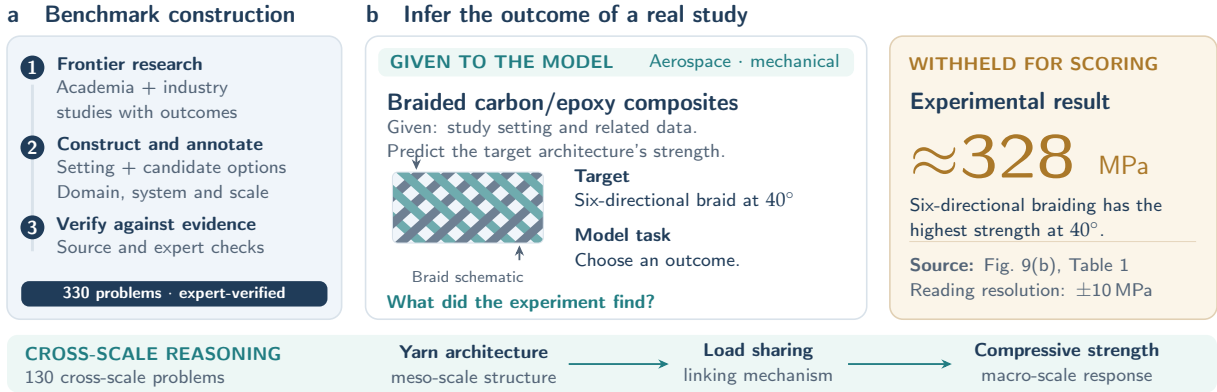
\begin{figure*}[t]
  \centering
  \begingroup
\definecolor{fmNavy}{HTML}{203C59}
\definecolor{fmTeal}{HTML}{217D82}
\definecolor{fmGold}{HTML}{AA7829}
\definecolor{fmSlate}{HTML}{536477}
\definecolor{fmLine}{HTML}{D9E2EA}
\definecolor{fmBlueBg}{HTML}{F0F5FA}
\definecolor{fmTealBg}{HTML}{EDF7F5}
\definecolor{fmGoldBg}{HTML}{FBF5E9}
\resizebox{\textwidth}{!}{%
\sffamily\fontsize{7.5}{9.4}\selectfont
\begin{tikzpicture}[x=1cm,y=1cm,
  font=\sffamily\fontsize{7.5}{9.4}\selectfont,
  text=fmNavy,
  txt/.style={anchor=north west,align=left,inner sep=0pt},
  panel/.style={rounded corners=5pt,draw=fmLine,line width=.6pt},
  flow/.style={-{Stealth[length=4pt,width=3.5pt]},draw=fmSlate,line width=.7pt},
  badge/.style={circle,fill=fmNavy,text=white,inner sep=0pt,minimum size=.32cm,font=\sffamily\bfseries\fontsize{7}{8}\selectfont}]
\path[use as bounding box] (0,0.12) rectangle (16.2,-5.23);
\node[txt,font=\sffamily\bfseries\fontsize{9}{11}\selectfont] at (0,0) {a\quad Benchmark construction};
\node[txt,font=\sffamily\bfseries\fontsize{9}{11}\selectfont] at (4.8,0) {b\quad Infer the outcome of a real study};
\draw[panel,fill=fmBlueBg] (0,-.43) rectangle (4.48,-4.22);
\draw[fmLine,line width=1.1pt] (.34,-1.08) -- (.34,-3.31);
\node[badge] at (.34,-.85) {1};
\node[txt,text width=3.51cm,font=\sffamily\bfseries] at (.66,-.70) {Frontier research};
\node[txt,text=fmSlate,text width=3.51cm] at (.66,-1.01)
  {Academia + industry\\studies with outcomes};
\node[badge] at (.34,-1.92) {2};
\node[txt,text width=3.51cm,font=\sffamily\bfseries] at (.66,-1.77) {Construct and annotate};
\node[txt,text=fmSlate,text width=3.51cm] at (.66,-2.08)
  {Setting + candidate options\\Domain, system and scale};
\node[badge] at (.34,-2.99) {3};
\node[txt,text width=3.51cm,font=\sffamily\bfseries] at (.66,-2.84) {Verify against evidence};
\node[txt,text=fmSlate,text width=3.51cm] at (.66,-3.15)
  {Source and expert checks};
\fill[fmNavy,rounded corners=3pt] (.19,-3.71) rectangle (4.29,-4.05);
\node[text=white,font=\sffamily\bfseries\fontsize{6.5}{8}\selectfont] at (2.24,-3.88)
  {\NumAll problems $\cdot$ expert-verified};
\draw[panel,fill=white] (4.80,-.43) rectangle (11.51,-4.22);
\fill[fmTealBg,rounded corners=3pt] (4.97,-.59) rectangle (11.34,-.97);
\node[txt,text=fmTeal,font=\sffamily\bfseries] at (5.12,-.68) {GIVEN TO THE MODEL};
\node[anchor=east,text=fmTeal,inner sep=0pt] at (11.16,-.78) {Aerospace $\cdot$ mechanical};
\node[txt,text width=6.10cm,font=\sffamily\bfseries\fontsize{9}{11}\selectfont] at (5.08,-1.17)
  {Braided carbon/epoxy composites};
\node[txt,text=fmSlate,text width=5.96cm] at (5.08,-1.54)
  {Given: study setting and related data.\\Predict the target architecture's strength.};
\begin{scope}
  \clip[rounded corners=2pt] (5.16,-2.25) rectangle (7.18,-3.20);
  \fill[fmBlueBg] (5.16,-2.25) rectangle (7.18,-3.20);
  \foreach \x in {4.3,4.7,...,7.5}{
    \draw[fmNavy!65,line width=3.3pt] (\x,-3.35) -- ++(1.45,1.3);
    \draw[fmTeal!65,line width=3.3pt] (\x,-2.08) -- ++(1.45,-1.3);
  }
\end{scope}
\draw[fmLine,rounded corners=2pt] (5.16,-2.25) rectangle (7.18,-3.20);
\draw[flow] (5.48,-2.03) -- (5.48,-2.22);
\draw[flow] (6.86,-3.43) -- (6.86,-3.23);
\node[anchor=north,text=fmSlate,font=\sffamily\fontsize{6.5}{8}\selectfont] at (6.17,-3.48) {Braid schematic};
\node[txt,text width=3.44cm] at (7.59,-2.20)
  {\textbf{Target}\\Six-directional braid at $40^\circ$\\[4pt]
   \textbf{Model task}\\Choose an outcome.};
\node[txt,text=fmTeal,font=\sffamily\bfseries] at (5.08,-3.89)
  {What did the experiment find?};
\draw[panel,fill=fmGoldBg,draw=fmGold!35] (11.82,-.43) rectangle (16.20,-4.22);
\node[txt,text width=3.80cm,text=fmGold,font=\sffamily\bfseries] at (12.08,-.70) {WITHHELD FOR SCORING};
\node[txt,text width=3.80cm,font=\sffamily\bfseries\fontsize{9}{11}\selectfont] at (12.08,-1.17) {Experimental result};
\node[anchor=west,inner sep=0pt,text=fmGold,font=\sffamily\bfseries\fontsize{23}{25}\selectfont] at (12.07,-2.04) {$\approx$328};
\node[anchor=west,inner sep=0pt,text=fmGold,font=\sffamily\fontsize{10}{12}\selectfont] at (14.60,-2.15) {MPa};
\node[txt,text width=3.8cm] at (12.08,-2.59)
  {Six-directional braiding has the\\highest strength at $40^\circ$.};
\draw[fmGold!30] (12.08,-3.18) -- (15.92,-3.18);
\node[txt,text=fmSlate,text width=3.8cm] at (12.08,-3.39)
  {\textbf{Source:} Fig.~9(b), Table~1\\Reading resolution: $\pm$10\,MPa};
\fill[fmTealBg,rounded corners=4pt] (0,-4.43) rectangle (16.2,-5.20);
\node[txt,text=fmTeal,font=\sffamily\bfseries] at (.23,-4.59) {CROSS-SCALE REASONING};
\node[txt,text=fmSlate] at (.23,-4.89) {\NumAllCross cross-scale problems};
\node[align=center] (structure) at (6.25,-4.82) {\textbf{Yarn architecture}\\\textcolor{fmSlate}{meso-scale structure}};
\node[align=center] (mechanism) at (10.06,-4.82) {\textbf{Load sharing}\\\textcolor{fmSlate}{linking mechanism}};
\node[align=center] (property) at (14.19,-4.82) {\textbf{Compressive strength}\\\textcolor{fmSlate}{macro-scale response}};
\draw[flow,draw=fmTeal] (7.52,-4.82) -- (8.86,-4.82);
\draw[flow,draw=fmTeal] (11.26,-4.82) -- (12.65,-4.82);
\end{tikzpicture}%
}
\endgroup
  \caption{Overview of \bench. \textbf{(a)} A unified pipeline constructs and verifies problems from frontier academic and industrial research. \textbf{(b)} An abridged example separates the study setting given to the model from the experimental result withheld for scoring. The schematic and reasoning chain illustrate how yarn architecture connects to macroscopic strength; full problem details appear in \cref{app:examples}.}
  \label{fig:overview}
\end{figure*}

We introduce \benchlogo\,\bench (\cref{fig:overview}), a benchmark for evaluating language models on frontier materials R\&D. Its \NumAll problems follow three design principles:
\begin{itemize}[leftmargin=*,itemsep=1pt,topsep=2pt]
  \item \textbf{Grounded in frontier research.} Problems draw on academic and industrial research in fields including aerospace, semiconductors, embodied intelligence and nuclear fusion. Leading experimentalists and industry experts in each field validated every problem as relevant to their work.
  \item \textbf{Research outcomes as answers.} Each problem asks for an experimental outcome or conclusion supported by a real study, requiring prediction from its setting or interpretation of evidence.
  \item \textbf{Explicit domain and scale annotations.} Problems are labelled by functional domain, material system and the scale of the governing mechanism: micro, meso or macro. The \NumAllCross cross-scale problems also record the reasoning chain.
\end{itemize}

Evaluating seven proprietary, open-weight and domain-adapted models, we find:
\begin{itemize}[leftmargin=*,itemsep=1pt,topsep=2pt]
  \item \textbf{Performance remains limited.} The strongest general-purpose model, \BestModel, scores \BestAcc\% (\cref{fig:leaderboard}), \NumSolvedNone problems are solved by no general-purpose model, and half are solved by at most one (\S\ref{sec:evaluation}, \S\ref{sec:analysis-scale}).
  \item \textbf{Cross-scale reasoning is difficult.} With answer format and construction channel held fixed, cross-scale problems are the hardest, \CrossGap points below single-scale problems (\S\ref{sec:analysis-scale}).
  \item \textbf{Models select unsupported outcomes.} \AddWrongShareErr\% of parsed, incorrect responses to multi-answer problems endorse an outcome the study did not report (\S\ref{sec:analysis-cases}).
\end{itemize}

We also use \bench to guide domain adaptation of Qwen3.5-35B-A3B through continued pretraining (CPT) and supervised fine-tuning (SFT) with two scale-aware components: CPT data weighted by its benefit to cross-scale problems, and scale-specific modules trained by self-distillation (\S\ref{sec:adaptation}). Adaptation raises its \bench score from \BaseAcc\% for the official post-trained release to \OursAcc\%, matching the strongest proprietary model, with gains at every scale and on public materials benchmarks and no decline in mean performance on the general benchmarks evaluated.

\paragraph{Contributions.}
\begin{enumerate}[leftmargin=*,itemsep=1pt,topsep=2pt]
  \item \bench, \NumAll problems from frontier materials research with research outcomes as answers and domain and scale annotations (\S\ref{sec:benchmark}).
  \item An evaluation of seven models analysing where difficulty lies and how models err (\S\ref{sec:evaluation}--\ref{sec:analysis}).
  \item A scale-aware adaptation recipe that adds \AdaptGain points to an open-weight model and transfers to materials benchmarks (\S\ref{sec:adaptation}).
\end{enumerate}

\section{Related Work}
\label{sec:related}

\paragraph{Problem sources.}
Materials benchmarks for language models primarily use educational material or problems generated from existing text. MaScQA~\citep{zaki2024mascqa} collects 650 problems from a graduate entrance examination, while MaterialBENCH~\citep{yoshitake2024materialbench} and MatSciBench~\citep{zhang2025matscibench} draw on university textbooks; MatSciBench uses a 31-subfield taxonomy. MSQA~\citep{cheung2025msqa} and SciKnowEval~\citep{feng2024sciknoweval} use language models to generate problems from papers, textbooks and databases. Beyond materials, ChemBench~\citep{mirza2025chembench} assesses chemical knowledge, reasoning and intuition, GPQA~\citep{rein2024gpqa} collects expert-written graduate-level questions, and SuperGPQA~\citep{map2025supergpqa} and MMLU-Pro~\citep{wang2024mmlupro} build harder multi-discipline problem sets. Other benchmarks target property prediction from composition or structure~\citep{dunn2020matbench,rubungo2024llm4mat}, the use of materials simulation tools~\citep{liu2025mattools} or scientific images~\citep{alampara2025macbench}.

\paragraph{Answer targets and scale.}
The problem-solving benchmarks above take their answers from textbooks, answer keys or source documents, and their tasks range from recall to calculation, reasoning and prediction. \bench centres on experimental outcomes and conclusions from active research, asking models to infer withheld results or interpret selected study evidence (\cref{tab:compare}). Existing benchmarks organise problems by subfield or format but do not annotate the scale of the mechanism that determines the answer.

\paragraph{Sources of difficulty.}
Benchmark difficulty may reflect the level of study, expert labels (ChemBench), tiers assigned from model behaviour (MatSciBench uses the response lengths of six models), or model-based screening (SuperGPQA). In \bench, difficulty comes from the research problems themselves. Following work on construct validity~\citep{bean2025measuring}, signal and noise in evaluation~\citep{heineman2025signal} and errors in benchmark labels~\citep{gema2025mmluredux}, we test whether its verdicts are stable and its problems discriminate (\S\ref{sec:reliability}).

\begin{table*}[t]
\centering\footnotesize
\setlength{\tabcolsep}{3pt}
\begin{tabular}{@{}l>{\raggedright\arraybackslash}p{2.3cm}>{\raggedright\arraybackslash}p{2.0cm}ccc>{\raggedright\arraybackslash}p{2.4cm}@{}}
\toprule
\textbf{Benchmark} & \textbf{Problem source} & \textbf{Answer is} & \makecell{\textbf{Frontier}\\\textbf{research}} & \makecell{\textbf{Scale}\\\textbf{labels}} & \makecell{\textbf{R\&D}\\\textbf{practitioners}} & \textbf{Difficulty from} \\
\midrule
\multicolumn{7}{@{}l}{\emph{Teaching material}} \\
\makecell[l]{MaScQA\\\citep{zaki2024mascqa}}                       & Entrance examination          & Textbook knowledge & \xmark & \xmark & \xmark & Exam level \\
\makecell[l]{MaterialBENCH\\\citep{yoshitake2024materialbench}}    & Two textbooks                 & Textbook knowledge & \xmark & \xmark & \xmark & Course level \\
\makecell[l]{MatSciBench\\\citep{zhang2025matscibench}}            & Ten textbooks                 & Textbook knowledge & \xmark & \xmark & \xmark & Tiers by LLM response length \\
\multicolumn{7}{@{}l}{\emph{Generated from text}} \\
\makecell[l]{MSQA\\\citep{cheung2025msqa}}                         & Papers (2000--2021)           & Methods and results in source papers & \xmark & \xmark & \xmark & -- \\
\makecell[l]{SciKnowEval\\\citep{feng2024sciknoweval}}             & Papers, textbooks, databases  & Knowledge, prediction and application tasks & \xmark & \xmark & \xmark & Five knowledge levels \\
\multicolumn{7}{@{}l}{\emph{Expert-curated, multi-domain}} \\
\makecell[l]{ChemBench\\\citep{mirza2025chembench}}                & Experts, exams, databases     & Knowledge, reasoning, intuition & \xmark & \xmark & \xmark & Basic / advanced labels \\
\makecell[l]{SuperGPQA\\\citep{map2025supergpqa}}                  & Expert-collected resources    & Graduate-level knowledge and reasoning & \xmark & \xmark & \xmark & Screening with LLMs \\
\midrule
\bench (ours)                                       & Frontier academic and industrial research   & What a study found & \cmark & \cmark & \cmark & Frontier research outcomes \\
\bottomrule
\end{tabular}
\caption{Comparison of \bench with related benchmarks. \textbf{Answer is}: what the correct answer draws on; \bench targets experimental outcomes and conclusions drawn from study evidence. \textbf{Frontier research}: problems come from current research rather than established curricula. \textbf{Scale labels}: each problem records the scale of the governing mechanism and whether it crosses scales. \textbf{R\&D practitioners}: researchers in active academic and industrial R\&D contributed research problems and assessed their relevance to current work; \xmark: not reported (expert authoring or review alone does not count). \textbf{Difficulty from}: how problem difficulty is established or controlled.}
\label{tab:compare}
\end{table*}

\paragraph{Domain adaptation of LLMs for science.}
Continued pretraining on in-domain text is an established route to specialisation~\citep{gururangan2020dont}. In materials science, MatSciBERT~\citep{gupta2022matscibert} pretrains an encoder on materials text and HoneyBee~\citep{song2023honeybee} applies progressive instruction tuning. Continual instruction tuning can cause catastrophic forgetting of general abilities~\citep{luo2023forgetting}, and on MSQA several domain-specialised models trail general-purpose ones~\citep{cheung2025msqa}. Our recipe builds on example reweighting by meta-gradients~\citep{ren2018learning}, domain-mixture optimisation~\citep{xie2023doremi}, bottleneck adapters~\citep{houlsby2019parameter} and distillation from privileged context~\citep{lopezpaz2016unifying,snell2022learning}; organising them by scale is our extension. It is evaluated on frontier problems rather than textbook tasks (\S\ref{sec:adaptation}).

\section{The \bench Benchmark}
\label{sec:benchmark}

\label{sec:design}
\bench follows the three principles in \S\ref{sec:intro}: it draws problems from frontier research, asks models to infer study outcomes, and annotates domain and scale. We describe construction in \S\ref{sec:construction} and coverage in \S\ref{sec:taxonomy}.

\subsection{Construction}
\label{sec:construction}
\paragraph{Mining frontier studies.}
We worked with academic laboratories and industrial R\&D teams at the research frontier to identify problems from their recent work, including results not yet published. We also drew on experimental studies in specialist journals across aerospace, semiconductors, embodied intelligence and nuclear fusion. These sources enter a common construction pipeline: each study must provide a measured outcome or a conclusion supported by data, with sufficient context to formulate a problem. Experts assess its relevance to current research, rather than publication date alone (\cref{app:review}).

\paragraph{Writing problems.}
Each problem presents a study setting and selected evidence, then asks the model to identify a target outcome or draw a conclusion from the evidence. Some problems require quantitative derivation rather than prediction of an unreported measurement. All use single-answer or multi-answer multiple-choice formats. Distractors differ from the supported result in direction, magnitude or mechanism. Annotators with graduate training in materials science follow shared guidelines and record the source evidence for each answer, withholding the answer and its evidence locator from the model (\cref{app:review}).

\paragraph{Verification and expert review.}
Verification proceeds in two stages. Two language models independently check each problem for consistency with its source. Leading experimentalists and industry experts then review every problem in their field for relevance to their work, significance of the withheld outcome, and whether distractors represent plausible mistakes by a competent researcher. Experts familiar with the originating work review problems drawn from partner research. Problems that failed any criterion were revised or removed (\cref{app:review}). The problem set, options and answers were frozen before any model evaluation.

\subsection{Domain and scale}
\label{sec:taxonomy}
\begin{figure*}[t]
  \centering
  \includegraphics[width=0.74\textwidth]{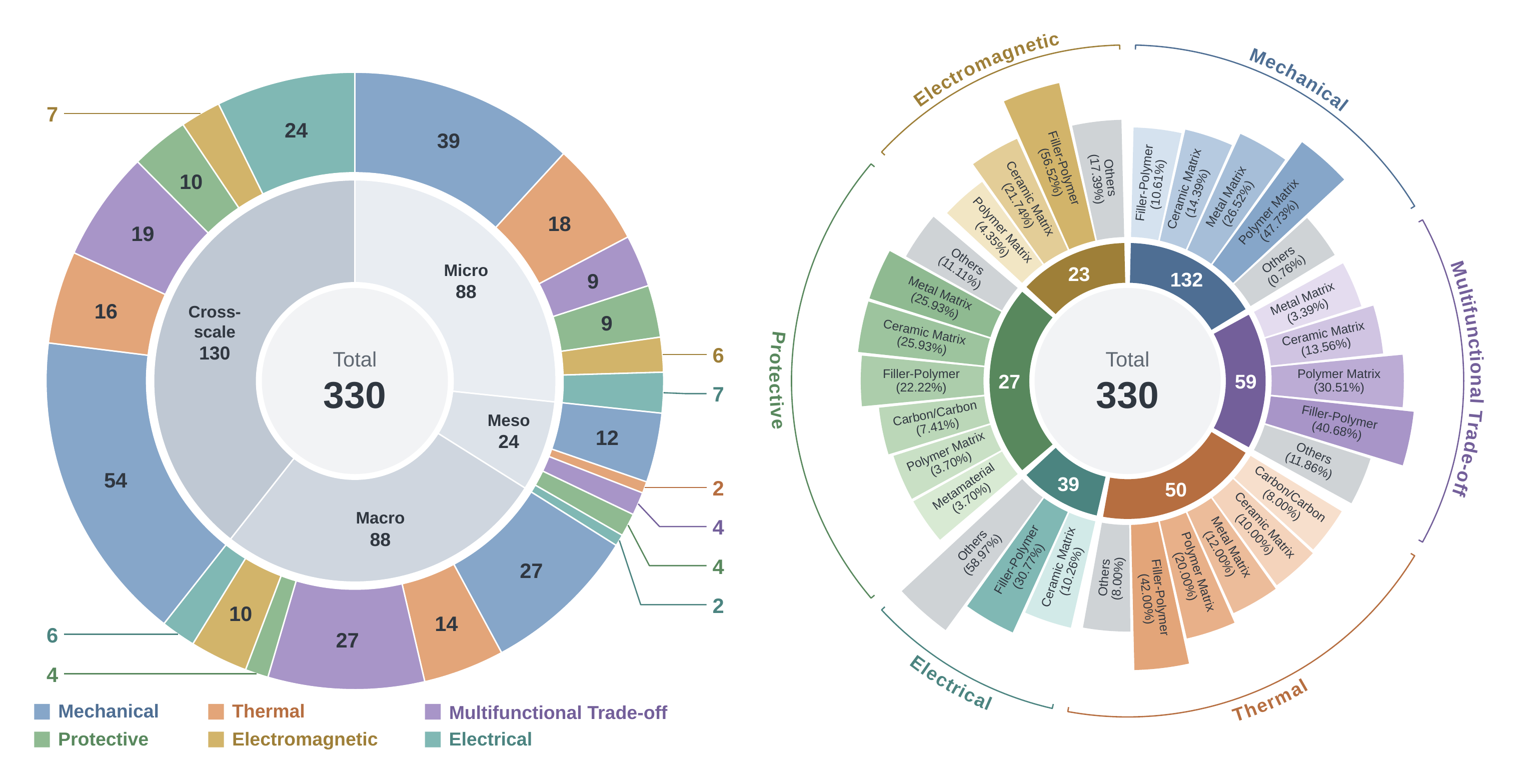}
  \caption{Composition of all \NumAll problems in \bench. \textbf{Left:} problems by scale of the governing mechanism (inner ring) and, within each scale, by functional domain (outer ring). \textbf{Right:} problems by functional domain (inner ring) and, within each domain, by material system (outer bars; percentages are shares within the domain).}
  \label{fig:distribution}
\end{figure*}

We annotate every problem by domain and scale to characterise its scientific context and reasoning demands (\cref{fig:distribution}).

\paragraph{Domain.}
The \emph{functional domain} identifies the property under investigation: \emph{mechanical} (strength, stiffness, fatigue, damage), \emph{thermal} (conduction, expansion, stability), \emph{electrical}, \emph{electromagnetic} (shielding, absorption, wave control), \emph{protective} (resistance to ablation, oxidation, irradiation and impact), and \emph{multifunctional trade-off}, which requires balancing two or more properties. The \emph{material system} identifies the material class: polymer-matrix, filler--polymer, ceramic-matrix, metal-matrix and carbon/carbon composites, metamaterials, and others. Mechanical problems form the largest domain (\DistMechShare\%). Matrix composites dominate this domain, while filler--polymer systems dominate electromagnetic and electrical problems.

\paragraph{Scale.}
Scale denotes the structural and mechanistic level that determines the answer, rather than a strict length scale. \emph{Micro} concerns atoms, lattices, defects and local structure; \emph{meso} concerns grains, phase domains, fibres or particles, pores and microstructure; \emph{macro} concerns specimens, devices and bulk response. A problem is \emph{cross-scale} when the answer requires linking levels, for example predicting a macroscopic property from micro- or mesostructure; we also annotate the reasoning chain (e.g., \emph{yarn architecture $\rightarrow$ load sharing $\rightarrow$ structural strength}). \bench contains \NumAllSingle single-scale problems (\NumAllMicro micro, \NumAllMeso meso, \NumAllMacro macro) and \NumAllCross cross-scale problems. All domain--scale combinations except meso-scale electromagnetic problems are represented (\cref{fig:distribution}, left), supporting analysis of performance across domains and scales (\S\ref{sec:analysis-scale}). Domain distributions vary across scales: multifunctional trade-offs are most numerous at the macro scale, and electrical problems are predominantly cross-scale (\DistElecCross of \DistElec). Labelling rules are given in \cref{app:format}.

Full statistics appear in \cref{tab:stats} (\cref{app:format}). Coverage varies across functional domains and scales, so we report results by domain and scale without claiming uniform coverage of the field.

\definecolor{fmRow}{HTML}{EDF7F5}
\providecommand{\bb}[1]{\textbf{#1}}
\providecommand{\ub}[1]{\underline{#1}}
\begin{table*}[t]
\centering\small
\setlength{\tabcolsep}{7pt}
\renewcommand{\arraystretch}{1.08}
\begin{tabular}{@{}llc cccc cc@{}}
\toprule
& & & \multicolumn{4}{c}{\textbf{Scale of the governing mechanism}} & \multicolumn{2}{c}{\textbf{Domain group}} \\
\cmidrule(lr){4-7}\cmidrule(l){8-9}
\textbf{Model} & \textbf{Reasoning} & \textbf{All} & Micro & Meso & Macro & Cross & Structural & Functional \\
\# problems & & \NumAll & \NumAllMicro & \NumAllMeso & \NumAllMacro & \NumAllCross & 159 & 171 \\
\midrule
\multicolumn{9}{@{}l}{\textit{Proprietary frontier models}} \\
GPT-6-Astra            & xhigh    & \ub{\BestAcc} & 51.1 & \ub{47.9} & \ub{39.2} & \ub{66.3} & \ub{55.1} & 52.3 \\
GPT-6-Sol              & xhigh    & \SolAcc  & 48.3 & \bb{50.0} & 34.1 & 65.4 & 50.0 & \ub{52.6} \\
Claude Opus 5.5        & max      & \OpusAcc & 43.2 & 35.4 & 27.3 & 65.8 & 47.8 & 46.8 \\
\addlinespace[2pt]
\multicolumn{9}{@{}l}{\textit{Open-weight frontier models}} \\
DeepSeek V4 Pro        & thinking & \DSAcc   & \ub{52.3} & 27.1 & 27.8 & 61.5 & 49.1 & 46.2 \\
GLM-5.3                & thinking & \GLMAcc  & 43.8 & 31.2 & 23.9 & 62.7 & 45.3 & 44.7 \\
\addlinespace[2pt]
\multicolumn{9}{@{}l}{\textit{Qwen3.5-35B-A3B (open-weight, 3B active parameters)}} \\
Official post-trained  & thinking & \BaseAcc & 30.3 & 22.0 & 23.8 & 52.4 & 35.7 & 37.6 \\
\rowcolor{fmRow}
\OursModel (ours)      & thinking & \bb{\OursAcc} & \bb{57.5} & 35.3 & \bb{41.6} & \bb{67.1} & \bb{57.3} & \bb{53.7} \\
\rowcolor{fmRow}
\quad\textit{gain from adaptation} & & \textit{+\AdaptGain} & \textit{+27.2} & \textit{+13.3} & \textit{+17.8} & \textit{+14.7} & \textit{+21.6} & \textit{+16.1} \\
\bottomrule
\end{tabular}
\caption{Score (\%) on \bench (scoring rule in \S\ref{sec:setup}). \textbf{Reasoning}: the highest reasoning setting each model offers. \textbf{Domain group}: structural problems (mechanical, protective) and functional problems (thermal, electrical, electromagnetic, multifunctional trade-off); per-domain scores in \cref{app:results}. The two Qwen rows are means over 16 samples with identical prompts and decoding (\S\ref{sec:adaptation}); cross-scale and single-scale columns differ in answer format and construction channel (\S\ref{sec:analysis-scale}). Bold: best per column; underlined: second best.}
\label{tab:main}
\end{table*}

\section{Evaluating Language Models}
\label{sec:evaluation}

\subsection{Setup}
\label{sec:setup}
\paragraph{Models.}
We evaluate seven models: three proprietary frontier models (GPT-6-Astra, GPT-6-Sol and Claude Opus 5.5), three open-weight models (DeepSeek V4 Pro, GLM-5.3 and the official post-trained Qwen3.5-35B-A3B), and \OursModel, our domain-adapted Qwen3.5-35B-A3B (\S\ref{sec:adaptation}). We enable reasoning mode where available. \OursModel and the official Qwen3.5-35B-A3B use identical prompts, decoding settings and a 32k-token output limit; their scores average 16 samples per problem. \cref{app:protocol} gives reasoning and decoding settings.

\paragraph{Protocol and scoring.}
Evaluation is closed-book: each model receives only the study setting, candidate options and required answer format. We withhold answers, sources and other problems, and disable tools unless stated otherwise. A response scores 1 if its predicted option set matches the gold set, 0.5 if it is a non-empty proper subset of the gold set, and 0 if it contains any wrong option; single-answer problems are therefore scored 1 or 0. Unparseable outputs score 0. We report mean scores as percentages.

\paragraph{Statistics.}
We report 95\% confidence intervals from 10{,}000 bootstrap resamples of problems and compare models using paired permutation tests. We report differences that are not significant at $p<0.05$ as unresolved.

\subsection{Main results}
\label{sec:main-results}

\paragraph{Performance on frontier problems.}
\begin{finding}
Performance on frontier problems remains limited: the strongest general-purpose model achieves barely half the available score on \bench, leaving substantial room for improvement.
\end{finding}
\noindent \BestModel scores \BestAcc\% (95\% CI \BestCILow--\BestCIHigh; \cref{tab:main}), significantly ahead of the open-weight DeepSeek V4 Pro (\DSAcc\%) and GLM-5.3 (\GLMAcc\%; paired permutation test, $p<0.01$). GPT-6-Sol (\SolAcc\%) does not differ significantly from \BestModel or DeepSeek V4 Pro, nor does Claude Opus 5.5 (\OpusAcc\%) differ from either open-weight model. Domain adaptation raises the official Qwen3.5-35B-A3B from \BaseAcc\% to \OursAcc\%, the highest overall score but within the CI of \BestModel.

\paragraph{Performance across domains and scales.}
No model exceeds 67.1\% in any scale column of \cref{tab:main}, and scores by functional domain range from 32.2\% to 65.2\% (\cref{app:results}). \OursModel leads on both structural (57.3\%) and functional problems (53.7\%). Because answer format and construction channel vary across scales, we compare matched problems in \S\ref{sec:analysis-scale}.

\subsection{Is \bench a reliable instrument?}
\label{sec:reliability}
We assess ranking stability and whether individual problems distinguish stronger from weaker models using per-problem scores (\cref{app:analysis}).

\begin{finding}
\bench yields stable rankings among the five general-purpose models analysed, and most informative problems have positive discrimination. With \ShareSolvedNone\% of problems unsolved by any general-purpose model, it offers headroom beyond saturated benchmarks.
\end{finding}

\paragraph{Rankings are stable.}
Resampling the problems 10{,}000 times tests ranking stability among five general-purpose models: \BestModel ranks first in \BootTopAstra\% of resamples, and the two GPT-6 models remain ahead of the other three in \BootTier\%.

\paragraph{Problems discriminate and leave headroom.}
For each of the \NumInformative problems with non-constant scores across six models, we correlate its scores with scores on the remaining problems. \ShareDiscPos\% have positive correlations (median \MedianDisc), indicating that stronger models generally perform better on these problems. \NumSolvedNone problems remain unsolved by every general-purpose model (\S\ref{sec:analysis-scale}), indicating room for improvement beyond a near-ceiling benchmark (MaScQA: \MaScQABase\% versus \MaScQAOurs\% for the two Qwen models that \bench separates by \AdaptGain points).

\section{What \bench Reveals}
\label{sec:analysis}
Aggregate scores do not explain which research problems models find difficult or why they fail. Using \bench's annotations and the per-problem responses of six models (\cref{app:analysis}), we ask where difficulty lies and how models fail.

\subsection{Expert validation}
\label{sec:analysis-relevance}
The first test of a benchmark that claims to sample the frontier is whether the people working at that frontier recognise its problems. Every \bench problem was reviewed by leading experimentalists and industry experts in its field, who assessed whether it poses a question their own research or development work faces, whether the withheld outcome is the kind of result their work turns on, and whether the wrong options are mistakes a competent colleague could make. Problems that failed any of these tests were revised or removed (\S\ref{sec:construction}; \cref{app:review}). The problems that remain are, in the judgement of practitioners, questions from their own work rather than exercises about it.

\subsection{Where does the difficulty lie?}
\label{sec:analysis-scale}
\begin{figure*}[t]
  \centering
  \includegraphics[width=\textwidth]{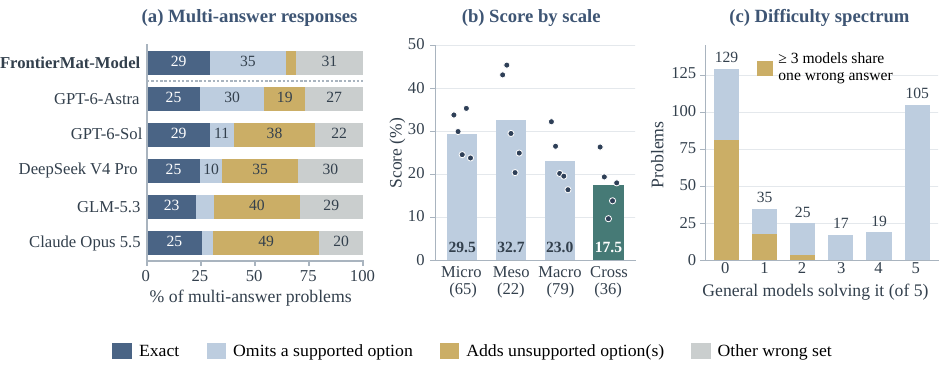}
  \caption{Where and how models fail on \bench. \textbf{(a)} Responses to the \NumMultiAns multi-answer problems. \textbf{(b)} Scores by scale on \NumStratum problems with answer format and construction channel held fixed (bars: mean of five general-purpose models, dots: individual models). \textbf{(c)} Problems by the number of general-purpose models solving them.}
  \label{fig:findings}
\end{figure*}

\begin{finding}
With answer format and construction channel held fixed, cross-scale problems are the most difficult. Performance is also lowest in multi-mechanism domains.
\end{finding}

\paragraph{Cross-scale problems are the hardest.}
The aggregate scores in \cref{tab:main} are higher for cross-scale problems (66.3\% versus 45.5\% for \BestModel), but the groups differ in composition: cross-scale problems are more often single-answer, and one construction channel contributes only cross-scale problems. Controlling for both factors reverses this pattern. Within the channel containing both scale types, the five general-purpose models average \StratMicro\% on micro, \StratMeso\% on meso and \StratMacro\% on macro problems, but only \StratCross\% on cross-scale problems among the \NumStratum multi-answer problems (\cref{fig:findings}b), lowest for \StratCrossLowestCount of five models. After weighting within-format differences by cross-scale problem counts (\cref{app:analysis}), the cross-scale score is \CrossGap points below single-scale problems (stratified permutation test, one-sided $p=\CrossGapP$), with model-specific gaps from \CrossGapMin points for GLM-5.3 to \CrossGapMax for DeepSeek V4 Pro. On the same matched problems, scores are lowest for electrical problems (\StratElec\%), which are mostly cross-scale because composite conductivity depends on filler dispersion and interfaces, and multifunctional trade-offs (\TradeoffStrat\%), which require balancing competing properties. These patterns point to difficulty linking scales and mechanisms.

\paragraph{Half of the problems are solved by at most one model.}
Difficulty varies substantially across problems (\cref{fig:findings}c). \NumSolvedAll problems are answered exactly by all five general-purpose models, while \NumSolvedAtMostOne (\ShareSolvedAtMostOne\%) are solved by at most one of them and \NumSolvedNone by none, with few problems between these extremes; on \NumSolvedNoneConsensus of the unsolved problems at least three models give the same wrong answer (gold in \cref{fig:findings}c). They span every scale and domain (\UnsolvedDomMin--\UnsolvedDomMax\% of each domain).

\subsection{How do models fail?}
\label{sec:analysis-cases}
\begin{finding}
Most errors include a plausible option unsupported by the study; the strongest general-purpose model makes this error least often.
\end{finding}

\paragraph{Over-endorsement dominates.}
Multi-answer problems require models to distinguish reported findings from other plausible outcomes. Of the general-purpose models' errors with a parsed answer on these problems, \AddWrongShareErr\% include at least one option the study did not support, and \OverShareErr\% contain every correct option together with at least one unsupported option (\cref{fig:findings}a). Claude Opus 5.5 selects \KOpus options on average, compared with \GoldK in the reference answer, and four of the five general-purpose models select strict supersets of the correct answer on \OverMinOthers--\OverMaxOthers\% of these problems. The corresponding rates are only \OverAstra\% for \BestModel and \OverOurs\% for \OursModel. \BestModel instead omits correct options on \UnderAstra\% of problems. Avoiding unsupported options distinguishes the strongest models.

\paragraph{Inferring the reported outcome.}
Model explanations illustrate both response patterns (\cref{app:cases}). On graphene-reinforced Cu--Cr--Zr alloys, \BestModel computes the Hall--Petch increment the study reports but declines to infer the withheld bulk strength: ``none of the proposed numerical comparisons is established''. That contribution alone does not establish bulk strength, making evidence sufficiency relevant to interpreting the response. On a carbon-fibre/polyamide problem, all three proprietary models explain why a stronger interface need not give a stronger composite, yet only GPT-6-Sol selects the measured outcome. On Cu-doped spinel thermistors, the model instead selects an activation-energy trend opposite to the measured one.

\section{Guiding Domain Adaptation with \bench}
\label{sec:adaptation}
\paragraph{Training.}
We adapt Qwen3.5-35B-A3B (3B of 35B parameters active). \emph{Continued pretraining} (CPT) uses 10.94B tokens, with a reference mixture of 78.5\% materials literature (2.13M papers on composites and two-dimensional materials), 1.5\% synthesised text and 20\% general replay. \emph{Supervised fine-tuning} (SFT) uses 133K instruction samples (23K materials-specific). No \bench problem, option or answer entered training (\cref{app:training}). On top of this recipe, two components target the capability \S\ref{sec:analysis-scale} found weakest: connecting evidence across scales.

\paragraph{Meta-weighted CPT.}
Natural sampling is dominated by well-studied systems and single-scale descriptions. \emph{Meta-weighted CPT} keeps the next-token objective but reweights materials data by their estimated utility for held-out cross-scale problems~\citep{ren2018learning,xie2023doremi}. Materials data are grouped into \NumAblBuckets buckets by material family, scale and source. A stronger model reads a data block as context and answers \NumAblProbe probe problems written from corpus papers; the block's value is the gain in answer margin over a word-shuffled copy of the same block, which controls for vocabulary and length. Standardised bucket values set the sampling weights once before CPT, with the token budget unchanged (\cref{app:training}).

\paragraph{Scale-modular self-distillation.}
Evidence at different scales comes from different sources and may conflict, so joint training lets one scale override another. After SFT, \emph{scale-modular self-distillation} attaches three low-rank residual modules~\citep{houlsby2019parameter}, one each for micro, meso and macro, to the final hidden state, and a gated fusion module combines their outputs before the shared output head. A frozen snapshot of the model reading the problem with the evidence of one scale teaches the corresponding module, and the snapshot reading all evidence teaches the fusion module~\citep{lopezpaz2016unifying,snell2022learning}. At inference the model sees only the problem, so each module must internalise the evidence it was taught with.

\paragraph{Results.}
Adaptation raises Qwen3.5-35B-A3B on \bench from \BaseAcc\% (official release) to \OursAcc\%, matching the strongest proprietary model and gaining at every scale and domain, from 13.3 points on meso to 27.2 points on micro problems, with 14.7 points on cross-scale problems (\cref{tab:main}, \cref{app:results}).

\begin{table}[t]
\centering\footnotesize
\setlength{\tabcolsep}{2.3pt}
\begin{tabular}{@{}lcccc@{}}
\toprule
& \multicolumn{3}{c}{\textbf{Materials}} & \textbf{General} \\
\cmidrule(lr){2-4}\cmidrule(l){5-5}
\textbf{Model} & \textbf{MSQA} & \textbf{MatSciB.} & \textbf{MaScQA} & \textbf{Know.} \\
\midrule
GPT-4 (CoT)        & --   & --   & \MaScQABest\unskip$^\star$ & -- \\
Gemini-2.0-Flash   & \MSQABest\unskip$^\star$ & -- & -- & -- \\
Claude-3.7-Sonnet  & 70.2 & --   & --   & -- \\
DeepSeek-R1        & --   & \textbf{\MSBBest\unskip}$^\star$ & -- & -- \\
GPT-5              & --   & 75.0 & --   & -- \\
\midrule
Qwen3.5-35B-A3B    & \MSQABase & \MSBBase & \MaScQABase & \GenAvgBase \\
\OursModel         & \textbf{\MSQAOurs} & \MSBOurs & \textbf{\MaScQAOurs} & \textbf{\GenAvgOurs} \\
\bottomrule
\end{tabular}
\caption{Public materials benchmarks and general ability (\%). Upper rows are reported by the original papers under their own settings, with $^\star$ for each paper's best. General Know.\ averages six benchmarks (\cref{tab:transfer}).}
\label{tab:transfer-main}
\end{table}

\paragraph{Transfer and cost.}
Under matched settings, adaptation improves MSQA by \MSQAGain points~\citep{cheung2025msqa} and MatSciBench by \MatSciBenchGain points~\citep{zhang2025matscibench} (\cref{tab:transfer-main}). Our scores exceed the original MSQA paper's best and fall \MSBGapBest points below MatSciBench's best, although those results use different settings. MaScQA scores are nearly unchanged. The six-benchmark general mean rises from \GenAvgBase to \GenAvgOurs, despite a \MMLUDrop-point drop on one benchmark: aggregate performance holds, with some task-level trade-offs.

\begin{table}[t]
\centering\small
\setlength{\tabcolsep}{5pt}
\begin{tabular}{@{}lcc@{}}
\toprule
\textbf{Qwen3.5-0.8B} & \textbf{Held-out} & \textbf{\bench} \\
\midrule
CPT + SFT (natural mixture) & \AblBaseHO & \AblBaseFM \\
+ meta-weighted CPT & \AblMetaHO\,{\scriptsize(+\AblMetaGainHO$^\dagger$)} & \AblMetaFM\,{\scriptsize(+\AblMetaGainFM)} \\
+ single-module self-distillation & \AblMixHO & \AblMixFM \\
+ scale-modular self-distillation & \AblSoEHO\,{\scriptsize(+\AblSoEGainHO$^\dagger$)} & \AblSoEFM\,{\scriptsize(+\AblSoEGainFM$^\dagger$)} \\
\bottomrule
\end{tabular}
\caption{Ablations on Qwen3.5-0.8B. Variants independently modify the natural-mixture CPT + SFT baseline. Held-out problems (\NumAblHeldOut) use exact match and \bench its standard score. Small figures are gains, and $^\dagger$ marks $p<0.05$ (paired bootstrap).}
\label{tab:ablation}
\end{table}

\paragraph{Component ablations.}
On Qwen3.5-0.8B with a 35B teacher (\cref{tab:ablation}, \cref{app:ablation}), meta-weighted CPT improves the CPT + SFT baseline by \AblMetaGainHO points on held-out problems and \AblMetaGainFM points on \bench. Scale-modular self-distillation adds \AblSoEGainHO and \AblSoEGainFM points, both significant, and scores above single-module self-distillation trained on the same distilled data.

\section{Conclusion}
\label{sec:conclusion}
Materials R\&D models must be evaluated at the research frontier. On \bench, the strongest general-purpose model scores \BestAcc\%, cross-scale reasoning remains difficult, and models select plausible but unsupported outcomes. Scale-aware adaptation guided by \bench brings an open-weight model level with the strongest proprietary model while maintaining mean performance on general benchmarks. We will extend \bench to open-ended answers and further material classes as the frontier moves.

\section*{Limitations}
\label{sec:limitations}

\paragraph{Size and coverage.} \bench contains \NumAll problems. Differences of a few points between models and estimates for small groups, such as meso-scale problems, are therefore subject to substantial uncertainty. Application fields and functional domains are unevenly represented, with composites over-represented among material systems. It therefore offers an uneven sample of frontier materials R\&D. We report results by domain and scale to make this variation explicit.

\paragraph{Reliance on source evidence.} Answers are taken from source studies and may reflect errors in the original findings. Some reported values are read from published plots, with the extraction resolution recorded for each problem. Language models were used to cross-check problems against their sources, which may favour problems that these models can readily parse.

\paragraph{Answer format.} Multiple-choice options provide information that may be unavailable in practical R\&D, and the number of correct options in multi-answer problems is unevenly distributed.

\paragraph{Adaptation study.} The study in \S\ref{sec:adaptation} evaluates adaptation on one main model, with component ablations on a smaller model from the same family, and its literature corpus covers mainly composites and two-dimensional materials. The findings may not generalise to other architectures, data mixtures or material systems, and the general-benchmark comparison in \cref{tab:transfer} covers only six benchmarks and cannot establish performance across all capabilities required of a general assistant.

\section*{Ethical Considerations}
\label{sec:ethics}
\paragraph{Expert participation.} The domain experts who nominated research directions and reviewed problems participated as research collaborators. Group identities remain anonymous during review.
\paragraph{Data and licensing.} Problems paraphrase research settings and reproduce a small number of reported values. Source evidence is recorded for every problem, and no figures or text are copied. \bench is intended for research evaluation.
\paragraph{Intended use.} \bench measures performance on multiple-choice problems derived from completed studies. A high score does not certify a model for unsupervised use in materials R\&D, where incorrect predictions can lead to costly experimental campaigns and compromise safety-critical components.
\paragraph{Use of AI assistants.} Language models were used to cross-check problems against their sources (\S\ref{sec:construction}) and to assist with writing and editing text and code. The authors reviewed all content and take responsibility for it.

\begingroup
\small
\bibliographystyle{acl_natbib}
\phantomsection
\bibliography{references}
\endgroup

\clearpage
\appendix
\section{Problem Format and Labelling Rules}
\label{app:format}

\begin{table}[h]
\centering\small
\begin{tabular}{@{}lr@{}}
\toprule
\textbf{Items} & \NumAll \\
\quad Single-answer / multi-answer & \NumAllSingleAns{} / \NumAllMultiAns \\
\textbf{Sources} & \\
\quad Partner R\&D groups (academic / industry) & \NumGroups{} (\NumAcademicGroups{} / \NumIndustryGroups) \\
\textbf{Scale of the governing mechanism} & \\
\quad Micro / meso / macro & \NumAllMicro{} / \NumAllMeso{} / \NumAllMacro \\
\quad Cross-scale & \NumAllCross \\
\bottomrule
\end{tabular}
\caption{Statistics of \bench. All items are multiple choice. Scale is the level of the structure and mechanism that governs the answer (\S\ref{sec:taxonomy}).}
\label{tab:stats}
\end{table}

\Cref{tab:stats} summarises \bench. Each problem contains the fields listed in \cref{tab:schema}. Fields marked $\dagger$ are hidden from models during evaluation.

\begin{table}[h]
\centering\footnotesize
\begin{tabular}{@{}lp{3.9cm}@{}}
\toprule
\textbf{Field} & \textbf{Content} \\
\midrule
\texttt{id} & Stable identifier \\
\texttt{prompt} & Study setting, related results, options, answer format \\
\texttt{answer\_format} & Single-answer or multi-answer \\
\texttt{gold}$^\dagger$ & Letter, or alphabetically ordered letter set \\
\texttt{domain} & Functional domain (\S\ref{sec:taxonomy}) \\
\texttt{material\_system} & Class of material (\S\ref{sec:taxonomy}) \\
\texttt{field} & Application field \\
\texttt{scale} & Micro, meso, macro or cross-scale \\
\texttt{scale\_chain}$^\dagger$ & For cross-scale problems, the chain of mechanisms across scales \\
\texttt{reference}$^\dagger$ & Reported result: values with units, or the categorical outcome \\
\texttt{evidence}$^\dagger$ & Figure, table and page of the result in the source \\
\texttt{source\_doi}$^\dagger$ & DOI or URL, where available \\
\texttt{source\_year} & Publication year, where available \\
\bottomrule
\end{tabular}
\caption{Schema of a problem record.}
\label{tab:schema}
\end{table}

\paragraph{Answer annotation.}
\begin{itemize}[leftmargin=*,itemsep=1pt]
  \item The answer is a result reported by the source study: a measurement, a quantity computed directly from measured data, or a conclusion supported by the study's data. Simulation outputs may provide context and are labelled as such, but cannot serve as answers to problems about experimental behaviour.
  \item Values read from plots are marked as such, together with the reading resolution (for example, about 10\,MPa for the problem in \cref{fig:overview}); numerical options differ by more than this resolution.
  \item If a source's summary statistic differs from its underlying values, the answer is recomputed from the values given in the prompt, and this decision is documented.
  \item Quantities use SI units or, where conventional in the field, the units used by the source. Units are stated in the prompt.
\end{itemize}

\paragraph{Scale annotation.}
The scale label identifies the structural or mechanistic level that determines the answer, rather than the specimen size. A problem is \emph{micro} if the answer depends on atoms, lattices, defects or local chemistry; \emph{meso} if it depends on grains, phases, fibres or particles, pores or their arrangement; \emph{macro} if it depends on the response of a specimen, component or device treated as a continuum. A problem is \emph{cross-scale} if answering it requires a mechanism linking information in the prompt at one level to an outcome at another. This chain is recorded as \emph{structure at one scale $\rightarrow$ mechanism $\rightarrow$ property at another}, for example \emph{clamping and temperature $\rightarrow$ hole-edge damage accumulation $\rightarrow$ joint fatigue life}. When two labels are plausible, annotators select the level of the mechanism that distinguishes the correct option from the strongest distractor.

\paragraph{Functional-domain annotation.}
The functional-domain label identifies the property addressed by the problem. A problem is labelled \emph{multifunctional trade-off} only if choosing the correct option requires weighing two or more properties against each other, not merely because several properties are mentioned.

\section{Construction and Expert Review}
\label{app:review}

\paragraph{Literature screening.}
We retrieved \NumLitCand journal articles published in 2022--2026. We first screened titles and abstracts for experimental materials studies in the four application fields, then reviewed the full texts of the \NumLitScreened remaining articles. Of these, we retained \NumLitPool as sources because they reported a specific experimental result with sufficient context to formulate a problem.

\paragraph{Research collaboration.}
\NumGroups R\&D groups participated (\NumAcademicGroups academic and \NumIndustryGroups industrial or national-laboratory). Their identities are withheld for anonymous review. They contributed problems grounded in their latest research, including work not yet published, and identified research directions for literature-based problem construction. Their work spans structural composites, ceramic components, polymer and coating systems, semiconductor defects, low-dimensional materials and electrocatalysts. Across sources, eligibility depends on experimental evidence, sufficient context and relevance to current R\&D.

\paragraph{Problem writing.}
Annotators followed the same guidelines across sources: state the study setting and selected evidence, ask for a measured outcome or a conclusion derived from that evidence, write plausible distractors, and record the basis for the answer. When the target is a withheld measurement, information that directly reveals it is removed. When the target requires quantitative interpretation, the necessary measurements are retained. Two language models checked each problem against its source (\S\ref{sec:construction}).

\paragraph{Expert review.}
For each problem, experts received the answer, the reported result and its location in the source, and an explanation of each option. They marked each problem Approve, Revise or Reject and provided free-text comments.

\paragraph{Source publication years.}
Published sources span \AllYearMin--\AllYearMax. Earlier studies are included for their relevance to problems identified by the participating researchers, rather than as evidence of recency.

\section{Evaluation Protocol}
\label{app:protocol}

\paragraph{Prompting.}
Models receive each problem in the format shown in \cref{app:examples}, in English, with no few-shot examples and no tools. Single-answer prompts end with ``\texttt{End with exactly: Answer: <letter>.}''; multi-answer prompts ask for all correct letters in alphabetical order.

\paragraph{Answer extraction and scoring.}
We parse the final ``\texttt{Answer:}'' line in each response, remove duplicate letters and sort the remaining letters. The parsed set receives 1 point if it matches the gold set, 0.5 points if it is a non-empty proper subset of the gold set, and 0 otherwise, including responses with an incorrect option and empty or unparseable answers.

\paragraph{Configurations.}
\Cref{tab:config} lists the reasoning and decoding settings of each model.
\begin{table}[h]
\centering\footnotesize
\begin{tabular}{@{}p{2.1cm}p{4.6cm}@{}}
\toprule
\textbf{Model} & \textbf{Setting} \\
\midrule
GPT-6-Astra & reasoning effort xhigh; median 71\,s, 2.2k output tokens \\
GPT-6-Sol & reasoning effort xhigh; median 39\,s, 1.4k tokens \\
Claude Opus 5.5 & extended thinking (max); median 121\,s, 5.0k tokens \\
DeepSeek V4 Pro & reasoning mode \\
GLM-5.3 & reasoning mode \\
Qwen3.5-35B-A3B & official post-trained release; same prompts and decoding as \OursModel; 16 samples \\
\OursModel & domain-adapted Qwen3.5-35B-A3B (\S\ref{sec:adaptation}); thinking mode, $T=0.6$, top-$p$ 0.95, at most 32k output tokens; 16 samples \\
\bottomrule
\end{tabular}
\caption{Model configurations.}
\label{tab:config}
\end{table}

\section{Full Results}
\label{app:results}
\Cref{tab:full} reports each model's score by functional domain.

\begin{table}[h]
\centering\footnotesize
\setlength{\tabcolsep}{2.5pt}
\resizebox{\colfigwidth}{!}{%
\begin{tabular}{@{}lccccccc@{}}
\toprule
\textbf{Subset} & \makecell{GPT-6\\Astra} & \makecell{GPT-6\\Sol} & \makecell{Claude\\Opus 5.5} & \makecell{DeepSeek\\V4 Pro} & \makecell{GLM-\\5.3} & \makecell{Qwen3.5\\(official)} & \makecell{\bench-\\Model} \\
\midrule
All & \BestAcc & \SolAcc & \OpusAcc & \DSAcc & \GLMAcc & \BaseAcc & \OursAcc \\
\midrule
Mechanical & 53.9 & 48.1 & 46.6 & 48.5 & 46.2 & 36.4 & 56.6 \\
Thermal & 50.7 & 50.0 & 48.0 & 51.0 & 47.0 & 34.4 & 49.8 \\
Multifunctional trade-off & 44.6 & 44.1 & 36.4 & 32.2 & 33.9 & 32.2 & 48.9 \\
Protective & 61.1 & 59.3 & 53.7 & 51.9 & 40.7 & 32.5 & 60.6 \\
Electromagnetic & 65.2 & 60.9 & 52.2 & 58.7 & 54.3 & 42.3 & 62.2 \\
Electrical & 58.5 & 64.1 & 57.7 & 53.8 & 52.6 & 46.9 & 61.0 \\
\bottomrule
\end{tabular}}
\caption{Scores (\%) by functional domain.}
\label{tab:full}
\end{table}

\section{Analysis Details}
\label{app:analysis}

\subsection{Scale analysis (\S\ref{sec:analysis-scale})}
All models are evaluated on the full set of \NumAll problems. For the controlled scale comparison, however, all problems contributed by partner groups are cross-scale, so that channel provides no within-channel single-scale comparison. We therefore use the \NumScaleComparison problems from the channel containing both scale types. The \NumStratum multi-answer problems within this set form the comparison in \cref{fig:findings}b. To estimate the overall scale gap, we use both answer formats within the \NumScaleComparison-problem set: we stratify by answer format (single-answer or multi-answer) and compute the difference in mean score between cross-scale and single-scale problems for each model and format. The reported gap averages these differences over formats and models, weighted by the number of cross-scale problems in each format. We permute scale labels within each format 5{,}000 times. The one-sided $p$-value is the proportion of permutations yielding a gap at least as negative as the observed gap.

\subsection{Per-problem analyses (\S\ref{sec:reliability}, \S\ref{sec:analysis})}
\paragraph{Answers.} Every analysis in \S\ref{sec:reliability} and \S\ref{sec:analysis} uses one answer per model and problem: the majority answer across samples where a model was sampled more than once, and the single answer otherwise. Scores computed from these answers therefore differ slightly from the means in \cref{tab:main}. The official Qwen3.5-35B-A3B is excluded from these analyses.

\paragraph{Response categories.} On a multi-answer problem with answer set $G$, a response $P$ is \emph{exact} if $P=G$, \emph{omits} if $P\subsetneq G$, \emph{adds} if $P\supsetneq G$, and \emph{other} otherwise (it both adds and omits, or is disjoint from $G$). An error contains an unsupported option if $P\not\subseteq G$.

\paragraph{Reliability.} For bootstrap rankings, we resample the \NumAll problems with replacement 10{,}000 times and rank the five general-purpose models by total score. For split-half agreement, we draw 2{,}000 random partitions into halves of 165 problems and report the mean Kendall $\tau$ between their rankings. Discrimination is the Pearson correlation, across the six models with per-problem answers, between a model's score on a problem and its total score on the other problems; it is undefined, and omitted, for the \NumConstant problems on which all six models receive the same score.

\section{Domain-Adaptation Details}
\label{app:training}

\subsection{CPT Corpus}
\paragraph{Sources.} Materials text comes from two internal literature collections: composites (about 2.03M records, 1.77M distinct papers after deduplication) and two-dimensional materials (1.40M records, 0.36M distinct papers). We add structured extractions (English and Chinese) and one-sentence summaries generated by Qwen3-32B (0.13B tokens), each prefixed with its source paper title and not manually checked, plus replay data from 23 public corpora.

\paragraph{Processing.} For each paper, we use the best available text, prioritising publisher XML (Elsevier, JATS), then PDF, then abstract. Re-parsing previously unparsed PDFs with PyMuPDF recovered about 400K papers. Cleaning removes reference lists, repeated headers and footers, page-number and copyright lines, line-break hyphenation and residual XML markup. Gopher-style heuristics on length, alphabetic ratio, duplicate lines and collapsed spacing remove low-quality documents. Exact deduplication uses the DOI or the normalised title and year, followed by cross-collection MinHash-LSH (128 permutations, Jaccard threshold 0.8). A fastText classifier (validation accuracy 0.936) labels documents as core ($p\geq0.8$), adjacent ($0.3\leq p<0.8$) or off-domain ($p<0.3$). We exclude documents outside English and Chinese, retracted papers and papers in clinical medicine, and hold out papers published in 2026 or later plus a random 1\%. The resulting corpus has 1.58M trainable documents and 10.60B tokens (Qwen3.5 tokenizer; 9.15B core, 0.25B adjacent, 1.21B off-domain), of which 8.34B come from the composites collection and 2.26B from the two-dimensional-materials collection. Documents are concatenated and cut into 4,096-token blocks; long documents are split, not truncated.

\paragraph{Buckets and packing.} Core and adjacent literature is grouped by document type (research papers versus reviews, books and chapters) and by quality tier (``hi'' if the paper's citation percentile within its publication year is at least 0.5, otherwise ``mid''). Documents are ordered by a salted hash, joined with end-of-sequence separators and cut continuously into 4,096-token blocks, so no text is truncated; block counts define natural sampling proportions. \Cref{tab:cpt-mix} reports pool sizes and the reference corpus mixture before materials data are reweighted.

\begin{table}[h]
\centering\footnotesize
\resizebox{\colfigwidth}{!}{%
\begin{tabular}{@{}lrr@{}}
\toprule
Bucket & Pool (B tok.) & Reference share (\%) \\
\midrule
Papers, hi & 4.11 & 40.0 \\
Papers, mid & 5.74 & 31.1 \\
Reviews, hi & 0.52 & 5.0 \\
Reviews, mid & 0.24 & 2.4 \\
Synthetic (extractions, summaries) & 0.13 & 1.5 \\
\midrule
English web / encyclopaedia & -- & 6.0 / 1.4 \\
Chinese web / encyclopaedia & -- & 3.0 / 0.6 \\
Source code / code solutions & -- & 3.0 / 1.0 \\
Math web / math solutions & -- & 2.75 / 1.25 \\
STEM textbooks & -- & 1.0 \\
\bottomrule
\end{tabular}}
\caption{CPT data pools and reference mixture for the 10.94B-token budget. Materials buckets are reweighted once before CPT using the in-context procedure below. These proportions describe the corpus mixture, not the learned weights within materials data. Papers are \emph{hi} if their citation count is at or above the median for their publication year and \emph{mid} otherwise (including papers without citation data); reviews, books and chapters form the review buckets. Replay (bottom, 20\% in total) draws on FineWeb-Edu, Ultra-FineWeb, FineWiki, Wikipedia, FineWeb2-HQ, Stack-Edu, StarCoderData, OpenCodeInstruct, CodeContests, OpenCodeReasoning, FineMath, InfiWebMath, OpenMathReasoning, NuminaMath-1.5, LibreTexts, Cosmopedia and Pressbooks.}
\label{tab:cpt-mix}
\end{table}

\paragraph{Replay share.} In small-model ablations, removing replay caused general capabilities, especially Chinese, mathematics and code, to deteriorate throughout training. A 10\% share was insufficient, whereas 15--25\% had little effect on held-out materials loss; we retain a 20\% replay share.

\subsection{SFT Data}
The final assistant turn in each sample contains either \texttt{<think>}reasoning\texttt{</think>} followed by the answer, or the answer with an empty reasoning block; earlier assistant turns retain only the answer. The dataset contains 133,101 training and 690 validation samples (304M tokens, 202M of them answer tokens); 39\% use reasoning and 39\% are in Chinese. \Cref{tab:sft-mix} lists the composition.

\begin{table}[h]
\centering\footnotesize
\setlength{\tabcolsep}{4pt}
\begin{tabular}{@{}p{2.4cm}rp{3.6cm}@{}}
\toprule
Part & Samples & Content \\
\midrule
General capability & 99.9K & mathematics, code and science reasoning (OpenThoughts3, LIMO, s1K, Codeforces-CoTs, Nemotron-Science); Chinese reasoning distillation; instruction following (Nemotron-IF, SmolTalk2); dialogue (Firefly, SmolTalk2); structured output \\
Materials tasks & 11.0K & extraction from papers (EN 3.6K, ZH 3.5K), summarisation 3.8K \\
General supplement & 9.9K & constrained instructions, rewriting, summarisation, Python, long context \\
Materials questions & 9.0K & single-choice 3.2K, multi-select 4.8K, calculation 1.0K \\
Format-aligned items & 3.4K & recomposed, true/false, numeric-variant and newly written items from an internal set disjoint from all evaluation sets \\
\bottomrule
\end{tabular}
\caption{SFT data composition.}
\label{tab:sft-mix}
\end{table}

\paragraph{Materials questions.} Qwen3-235B-A22B-Thinking generates single-choice, multi-select and calculation problems from corpus papers, with a true/false judgment and supporting evidence for every option. The same model then answers each problem without the paper. If its answer differs from the gold answer, it reviews the problem with the paper; problems whose gold answer remains unconfirmed are discarded. Reasoning samples are obtained by rejection sampling, retaining a closed-book chain of thought only if its final answer exactly matches the gold answer; non-reasoning samples are expert-style explanations that assess each option and end with the answer letters.

\subsection{Scale-Aware Training}
\paragraph{Meta-weighted CPT.} The objective is to favour materials data that improve performance on held-out cross-scale problems. A direct implementation would evaluate candidate bucket weights through virtual parameter updates and differentiate the cross-scale loss through those updates. We instead estimate data utility offline using a stronger model, then hold the resulting sampling weights fixed throughout CPT.

\paragraph{In-context approximation.} We use in-context learning as a proxy for the benefit of training on a data block, scoring buckets once before CPT. A stronger model reads a data block $d$ as reference material and answers each probe problem $j$; its margin $\Delta_j$ is the log-probability of the correct answer minus that of the strongest distractor. The value of $d$ is
\[
v(d)=\tfrac{1}{J}\textstyle\sum_j\big[\Delta_j(d)-\Delta_j(\tilde d)\big],
\]
where $\tilde d$ is $d$ with its words shuffled, providing a control with the same words and length but disrupted structure. The \NumAblProbe probe problems are written from corpus papers with a domain mix matching \bench's and contain no \bench problem. Materials data are grouped into \NumAblBuckets buckets by material family, scale and source; $z_b$ is the standardised mean value of bucket $b$, and materials data are resampled with $p_b\propto p_b^{\text{nat}}\exp(\lambda z_b)$ ($\lambda=3$), leaving the other data and the token budget unchanged.

\paragraph{Scale-modular self-distillation.} Let $h_t$ be the final hidden state for the problem $x$ and answer prefix $y_{<t}$. Each scale module $A_k$, $k\in\{\mu,m,M\}$, is a low-rank residual bottleneck, $z_k=h_t+A_k(h_t)$; the fusion module computes gates from $[h_t;z_\mu;z_m;z_M]$ and returns $\tilde h_t=h_t+\sum_k g_k\,(z_k-h_t)$, which the shared output head maps to the next-token distribution. Let $p_{\bar\theta}$ be a frozen snapshot of the model and $E_k$ the evidence for scale $k$. Define the teacher distributions
\[
\begin{aligned}
q_{k,t} &= p_{\bar\theta}(\cdot\mid x,E_k,y_{<t}),\\
q_{\mathrm{all},t} &= p_{\bar\theta}(\cdot\mid x,E_\mu,E_m,E_M,y_{<t}).
\end{aligned}
\]
The distillation loss at token $t$ is
\[
\begin{aligned}
\mathcal{L}_{\mathrm{dist},t}
&= \sum_k \mathrm{KL}\big(q_{k,t}\,\|\,p(\cdot\mid z_k)\big)\\
&\quad + \mathrm{KL}\big(q_{\mathrm{all},t}\,\|\,p(\cdot\mid\tilde h_t)\big).
\end{aligned}
\]
Training minimises this loss together with cross-entropy on the reference answer. Micro evidence describes defects, dislocations and interaction mechanisms; meso evidence describes grains, precipitates, fibre architecture, porosity and phase distribution; macro evidence summarises property curves and test and service conditions.

\subsection{Component Ablations at 0.8B}
\label{app:ablation}
\paragraph{Setup.} For the CPT comparison in \cref{tab:ablation}, we train Qwen3.5-0.8B with the same 1.2B-token budget and SFT procedure, using either natural sampling proportions or the in-context bucket weights above. Distillation variants start from the natural-mixture CPT + SFT model. \emph{Scale-modular self-distillation} is implemented with low-rank adapters on the frozen SFT model. Because a 0.8B model reads long evidence unreliably, the evidence-reading teacher in this ablation is a 35B model rather than a snapshot of the student; it answers 1{,}044 training problems with evidence of one scale, or with all scales plus the cross-scale causal statements reported by the authors, extracted from about 1{,}600 source papers, and only correct responses, including their reasoning, are kept as targets. A shared adapter trained on all distilled data is merged first; three scale adapters (rank 16) are then trained on the data of their scale, and a per-layer gate over them plus a fusion adapter on the all-evidence data. The \emph{single-module self-distillation} baseline trains a single rank-64 adapter on exactly the same distilled data.

\paragraph{Evaluation.} Reasoning is enabled with temperature 0.6. Held-out problems are generated as in \cref{app:training} from papers excluded from all training, sampled four times with up to 12k output tokens and scored by exact match. \bench is sampled eight times with up to 32k output tokens, over three seeds for the CPT comparison and two for the distillation comparison, and scored with its standard rule (\S\ref{sec:setup}). Differences use a paired bootstrap over problems. The two evaluations differ in metric and output budget, so the magnitudes of their gains are not directly comparable. \Cref{tab:ablation-full} lists all differences; meta-weighted CPT also lowers the held-out CPT materials loss from \AblLossBase to \AblLossMeta.

\begin{table}[h]
\centering\footnotesize
\setlength{\tabcolsep}{3pt}
\begin{tabular}{@{}p{2.6cm}cc@{}}
\toprule
\textbf{Comparison} & \textbf{Held-out} & \textbf{\bench} \\
\midrule
Meta-weighted vs.\ natural CPT & +\AblMetaGainHO\ \AblMetaCIHO & +\AblMetaGainFM\ \AblMetaCIFM \\
Scale-modular vs.\ CPT + SFT & +\AblSoEGainHO\ \AblSoECIHO & +\AblSoEGainFM\ \AblSoECIFM \\
Scale-modular vs.\ single-module & +\AblVsMixHO\ \AblVsMixCIHO & +\AblVsMixFM\ \AblVsMixCIFM \\
\bottomrule
\end{tabular}
\caption{Differences (points) in the 0.8B ablations, with 95\% paired-bootstrap intervals (held-out: exact match; \bench: score).}
\label{tab:ablation-full}
\end{table}

\subsection{Transfer and General-Benchmark Evaluation}
\paragraph{Truncation.} \BaseTrunc\% of the official model's responses on multi-answer problems reach the 32k-token limit; scoring all of them as correct would add at most \BaseTrunc points, compared with a gap of 22.2 points on those problems (\BaseFactAcc\% versus \OursFactAcc\%).

Both models in \cref{tab:transfer-main} and \cref{tab:transfer} are evaluated with the same harness, prompts and decoding settings, with reasoning mode enabled throughout. Materials benchmarks are scored with each benchmark's released protocol: the 1{,}757 binary questions of MSQA, the 1{,}025 text-only questions of MatSciBench and all 650 questions of MaScQA, the subsets for which the original papers report the results quoted in \cref{tab:transfer-main}. General benchmarks are run with lm-eval-harness on vLLM.
\begin{table}[h]
\centering\small
\setlength{\tabcolsep}{4pt}
\begin{tabular}{@{}lcc@{}}
\toprule
& \makecell{Qwen3.5-\\35B-A3B} & \OursModel \\
\midrule
\multicolumn{3}{@{}l}{\emph{Public materials benchmarks}} \\
MSQA (binary) & \MSQABase & \textbf{\MSQAOurs} \\
MatSciBench (text) & \MSBBase & \textbf{\MSBOurs} \\
MaScQA & \MaScQABase & \textbf{\MaScQAOurs} \\
\midrule
\multicolumn{3}{@{}l}{\emph{General benchmarks}} \\
MMLU & \textbf{84.8} & 82.5 \\
C-Eval & 84.6 & \textbf{85.5} \\
GSM8K & 84.6 & \textbf{85.9} \\
ARC-Challenge & 62.5 & \textbf{66.9} \\
WinoGrande & 74.6 & \textbf{74.7} \\
PIQA & 83.0 & \textbf{83.1} \\
Average (6) & 79.0 & \textbf{79.8} \\
\bottomrule
\end{tabular}
\caption{Scores (\%) of the official post-trained Qwen3.5-35B-A3B and \OursModel on public materials benchmarks and on six general benchmarks, under identical evaluation settings.}
\label{tab:transfer}
\end{table}

\subsection{Hyperparameters}
\begin{table}[h]
\centering\footnotesize
\begin{tabular}{@{}lp{3.6cm}@{}}
\toprule
& \textbf{CPT} \\
\midrule
Data & 10.94B tokens (\cref{tab:cpt-mix}) \\
Learning rate & $10^{-5}$, cosine to $10^{-6}$ \\
Batch, length & 256 $\times$ 4,096 \\
Steps & 10,433 \\
\bottomrule
\end{tabular}
\caption{Hyperparameters for CPT.}
\label{tab:training}
\end{table}

\section{Example Problems}
\label{app:examples}
The following examples reproduce the problems shown to models, followed by their hidden annotations.

\begin{examplebox}{Performance forecasting (CWB7-001, single)}
\footnotesize
Single-choice: select the ONE option closest to the held-out experimental outcome. Numerical options are approximate outcome hypotheses, not additional measurements.

\smallskip
Forecast axial compressive strength for an untested architecture--angle combination of RTM T700 carbon/TDE-86 epoxy. Four-directional braiding is the basic architecture. Five-directional adds longitudinal straight yarns; six-directional additionally adds transverse yarns; seven-directional additionally adds through-thickness yarns. Coupons are untrimmed; compression uses ASTM D6641 at 1 mm/min. The following five-specimen means are rounded graph readings. Each row is [directions, angle\_deg, fiber\_volume\_percent, strength\_MPa]:\newline
[4,20,57.7,447]; [5,20,60.4,507]; [6,20,56.9,496]; [7,20,50.3,313]; [4,40,55.2,220]; [5,40,51.0,290]; [7,40,56.0,226].\newline
Target: six-directional, 40 degrees, fiber volume 55.5\%, thickness 3.36 mm. Estimate the held-out measured mean; no deterministic constitutive law is stipulated.

\smallskip
A. About 330 MPa; 6d is strongest at 40 degrees.\newline
B. About 250 MPa; 5d is strongest at 40 degrees.\newline
C. About 190 MPa; 5d is strongest at 40 degrees.\newline
D. About 420 MPa; 6d is strongest at 40 degrees.

\smallskip
Give brief reasoning. End with exactly: Answer: <letter>.

\smallskip\hrule\smallskip
\textbf{Gold:} A \quad \textbf{Headline model:} B \quad \textbf{Measured reference:} \texttt{\{"strength\_MPa": 328, "strongest\_at\_40": "6d"\}}\\
\textbf{Evidence locator:} Figure 9(b), Table 1 \quad \textbf{Source:} 2022, en, expert-authored
\end{examplebox}

\begin{examplebox}{Data interpretation (CWB7-004, multi)}
\footnotesize
Multiple-select: select ALL statements consistent with the stated experiment and its evidence. More than one option is correct.

\smallskip
A matched T300/5405 laminate series underwent one common wet/hot aging program. Unaged$\rightarrow$aged measurements are:\newline
plain tension: failure load 40.39$\rightarrow$37.30 kN, modulus 54.10$\rightarrow$53.40 GPa;\newline
open-hole tension: 33.50$\rightarrow$28.94 kN, 56.15$\rightarrow$54.02 GPa;\newline
undamaged compression: 78.05$\rightarrow$72.93 kN, 55.00$\rightarrow$53.13 GPa;\newline
15 J impact followed by compression: 59.20$\rightarrow$42.91 kN, 47.96$\rightarrow$46.20 GPa.\newline
Within each pair geometry is unchanged.

\smallskip
A. The impacted-compression load retention is about 72.5\%.\newline
B. Impacted compression has the largest relative failure-load loss among the four configurations.\newline
C. Open-hole tension retains less failure load, proportionally, than impacted compression.\newline
D. The impacted-compression modulus loss is about 27.5\%.

\smallskip
Give brief reasoning. End with exactly: Answer: <letters in alphabetical order>.

\smallskip\hrule\smallskip
\textbf{Gold:} AB \quad \textbf{Headline model:} AB \quad \textbf{Measured reference:} \texttt{\{"cai\_load\_retention\_pct": 72.483, "most\_sensitive": "impacted\_compression", "stiffness\_proxy\_valid": "no"\}}\\
\textbf{Evidence locator:} Tables 1--2, scanned pp437--438 \quad \textbf{Source:} 1998, zh, expert-authored
\end{examplebox}

\begin{examplebox}{Evidence judgment (CWB7-007, multi)}
\footnotesize
Multiple-select: select ALL statements consistent with the stated experiment and its evidence. More than one option is correct.

\smallskip
Through-thickness Darcy tests use 17 plies of SW220C100b glass satin and 100 cSt silicone oil. Fiber volume fractions span 46--54\%; injection pressures span .05--.30 MPa. A support plate has 58\% open area; adding a 40-mesh screen introduces an 82\%-open spreading layer. The measured fabric imprint shows that the effective flowing area shrinks as pressure increases. K is inferred from Q*viscosity*thickness/(area*pressure\_drop).\newline
Predict the observed joint outcome using categorical fields: with\_screen\_K\_vs\_pressure and without\_screen\_K\_vs\_pressure at 46\% fiber volume (each approximately\_constant/increases/decreases), K\_vs\_fiber\_fraction (increases/decreases), and screen\_effect\_on\_scatter (reduces/increases). Predictions refer to thickness- and effective-area-corrected permeability, not uncorrected flow rate.

\smallskip
A. Adding the screen increases between-test scatter.\newline
B. With the screen, corrected permeability remains approximately constant as pressure increases at 46\% fiber volume.\newline
C. Without the screen, corrected permeability increases as pressure increases at 46\% fiber volume.\newline
D. Higher fiber volume fraction lowers corrected permeability in both fixture configurations.

\smallskip
Give brief reasoning. End with exactly: Answer: <letters in alphabetical order>.

\smallskip\hrule\smallskip
\textbf{Gold:} BD \quad \textbf{Headline model:} BCD \quad \textbf{Measured reference:} \texttt{\{"with\_screen\_K\_vs\_pressure": "approximately\_constant", "without\_screen\_K\_vs\_pressure": "decreases", "K\_vs\_fiber\_fraction": "decreases", "screen\_effect\_on\_scatter": "reduces"\}}\\
\textbf{Evidence locator:} Figures 8--12 and conclusions, PDF pp7--8 \quad \textbf{Source:} 2020, zh, expert-authored
\end{examplebox}

\section{Case-Study Problems}
\label{app:cases}
We reproduce the three problems discussed in \S\ref{sec:analysis-cases}, as shown to models, with answers and supporting evidence. Each problem records its source study.

\begin{examplebox}{Case 1 (FMB-F125): cross-scale, micro $\rightarrow$ macro; metal-matrix composite, mechanical}
\footnotesize\sloppy
CeO$_2$/GO-Cu30Cr0.2Zr electrical contact composites are fabricated by hydrothermal synthesis and spark plasma sintering at 950\,$^\circ$C under 50\,MPa. Electron backscatter diffraction gives average grain sizes of 1.55\,\textmu m for Cu30Cr0.2Zr and 0.86\,\textmu m for 0.5CeO$_2$/GO-Cu30Cr0.2Zr. The Hall--Petch coefficient for Cu is approximately 0.071\,MPa\,m$^{1/2}$. After tensile deformation, Williamson--Hall analysis yields dislocation densities of $4.14\times10^{14}$\,m$^{-2}$ for Cu30Cr0.2Zr, $8.04\times10^{14}$\,m$^{-2}$ for 0.3CeO$_2$/GO-Cu30Cr0.2Zr, and $9.15\times10^{14}$\,m$^{-2}$ for 0.5CeO$_2$/GO-Cu30Cr0.2Zr. The Burgers vector for Cu is 0.256\,nm and the shear modulus is 42.1\,GPa. Uniaxial tensile testing at a constant deformation rate of 1\,mm/min gives a tensile strength of 276\,MPa for Cu30Cr0.2Zr and 367\,MPa for 0.3CeO$_2$/GO-Cu30Cr0.2Zr. The tensile strength of 0.5CeO$_2$/GO-Cu30Cr0.2Zr is withheld from the data above.

Which of the following statements hold?

A. The tensile strength of the 0.5\,wt\% composite is within 25\,MPa of the 367\,MPa measured for the 0.3\,wt\% composite.\\
B. The tensile strength of 0.5CeO$_2$/GO-Cu30Cr0.2Zr exceeds 390\,MPa.\\
C. The tensile strength increment from 0.3 to 0.5\,wt\% CeO$_2$/GO is larger than the increment from 0 to 0.3\,wt\% CeO$_2$/GO.\\
D. CeO$_2$ nanoparticles riveted on the graphene surface resist agglomeration caused by van der Waals forces, improving the dispersion of graphene in the Cu matrix.\\
E. Load redistribution from the graphene reinforcement is identified as the most important strengthening mechanism in CeO$_2$/GO-Cu30Cr0.2Zr composites.\\
F. The conductivity improvement in CeO$_2$/GO-Cu30Cr0.2Zr is partly attributed to the oxidation of graphene oxide during sintering, which partially recovers the intrinsic electrical conductivity of graphene.

\smallskip
\textbf{Answer: B, D, E.} The withheld strength is 406\,MPa, so B is correct and A and C are incorrect (the increment falls from 91 to 39\,MPa). The study computes the Hall--Petch contribution as 19.4\,MPa and describes it as negligible, identifying load redistribution to graphene as the dominant mechanism (E). F reverses the actual transformation: XPS shows thermal reduction, not oxidation, of graphene oxide.\\
\textbf{GPT-6-Astra selected D} (score 0.5), computing the Hall--Petch increment as 19.5\,MPa and declining to infer the strength; it rejected E as ``unsupported''.
\end{examplebox}

\begin{examplebox}{Case 2 (FMB-F071): interface $\rightarrow$ bulk; polymer-matrix composite, mechanical}
\footnotesize\sloppy
Short carbon fiber (10\,wt\%) reinforced semi-aromatic polyamide 6T (PA6T) composites were prepared using HF40S carbon fibers with different surface treatments. Pristine carbon fibers were coated with PA6T oligomer from 1\,wt\% and 2\,wt\% ethanol suspensions and then polymerized at 320\,$^\circ$C for 10\,min under vacuum, yielding CF-1 and CF-2 respectively. A separate sample, CF-1-PRE, used the 1\,wt\% coating heated only to 200\,$^\circ$C without polymerization.

For single-fiber microbond testing, an i-CF specimen was prepared by applying PA6T oligomer matrix droplets directly on a pristine fiber and polymerizing at 320\,$^\circ$C for 10\,min. For bulk composite testing, the i-CF composite was manufactured by in-situ polymerization of the entire PA6T oligomer and carbon fiber system under vacuum at 320\,$^\circ$C, so that the whole composite underwent the polymerization process rather than only a surface coating.

Microbond testing gave the following interfacial shear strength (IFSS) values: pristine CF $96\pm9$\,MPa, CF-1-PRE $93.8\pm12.4$\,MPa, CF-1 $107\pm13.6$\,MPa, CF-2 $117.6\pm10.7$\,MPa, and i-CF $130\pm18$\,MPa. AFM measurements showed the interphase thickness was approximately 350\,nm for pristine CF composites and 800\,nm for i-CF composites.

Tensile strength of the bulk composites was: pristine CF $184.5\pm17.9$\,MPa, CF-1-PRE $214.8\pm12.2$\,MPa, CF-1 $219.1\pm10.6$\,MPa, and CF-2 $223.1\pm9.9$\,MPa. The tensile strength of the i-CF bulk composite is withheld from the data above.

Which of the following statements hold?

A. The tensile strength of the i-CF composite is below 220\,MPa.\\
B. The high IFSS of i-CF results from covalent C--C bonds grafted between the PA6T chains and the carbon fiber surface during in-situ polymerization, producing adhesive failure at the fiber--matrix interface.\\
C. The tensile strength of i-CF composites is higher than that of CF-2 composites, consistent with the interfacial shear strength ranking in which i-CF exceeds CF-2.\\
D. The tensile strength of the i-CF composite is within 10\,MPa of the 223.1\,MPa tensile strength of CF-2 composites.

\smallskip
\textbf{Answer: A, D.} The withheld strength is 214.2\,MPa: below 220 (A) and 8.9\,MPa below CF-2 (D), so the IFSS ranking does not hold for bulk strength (C fails). The study attributes the strong interface to physical entanglement and hydrogen bonding with cohesive failure in the matrix, not to covalent grafting (B fails), and attributes this discrepancy to voids introduced when the whole composite is polymerised under vacuum above 300\,$^\circ$C.\\
\textbf{GPT-6-Sol selected A, D} (score 1); \textbf{GPT-6-Astra and Claude Opus 5.5 selected A} (score 0.5), stating that D ``requires the withheld i-CF tensile measurement''.
\end{examplebox}

\begin{examplebox}{Case 3 (FMB-F253): single-scale, meso; ceramic, electrical}
\footnotesize\sloppy
Ni$_{0.6-x}$Mn$_{1.8}$Co$_{0.6}$Cu$_x$O$_4$ ($0\le x\le0.25$) spinel NTC thermistor ceramics are synthesised by solid-state reaction and sintered at 1150\,$^\circ$C for 3\,h. All compositions reach a relative density above 97\%. X-ray diffraction shows that the undoped ceramic ($x=0$) is a pure cubic spinel, while the Cu-doped ceramics develop a tetragonal spinel phase whose relative content reaches 35.2\% at $x=0.25$. X-ray photoemission O\,1s spectra of the $x=0.15$ and $x=0.25$ ceramics are deconvoluted into lattice-oxygen and oxygen-vacancy-related components. Room-temperature resistivity $\rho_{25}$ falls from 271\,$\Omega$\,cm at $x=0$ to 20\,$\Omega$\,cm at $x=0.25$. AC impedance spectra of the $x=0.15$ and $x=0.25$ ceramics are described by an equivalent circuit that separates a grain resistance $R_g$ from a grain-boundary resistance $R_{gb}$, and the grain and grain-boundary activation energies $E_a(R_g)$ and $E_a(R_{gb})$ are obtained by Arrhenius fitting of those fitted resistances. SEM micrographs give average grain sizes of 3.22\,\textmu m at $x=0$, 3.55\,\textmu m at $x=0.1$, 3.78\,\textmu m at $x=0.15$ and 14.57\,\textmu m at $x=0.25$. The average grain size, in \textmu m, at $x=0.2$ is withheld from the data above.

Which of the following statements hold?

A. The increase in tetragonal spinel phase fraction with Cu doping is primarily caused by Jahn--Teller distortion of Cu$^{2+}$ ions at octahedral B sites.\\
B. The average grain size is more than 4\,\textmu m smaller than the 14.57\,\textmu m measured at $x=0.25$.\\
C. The relative contribution of the oxygen-vacancy-related O\,1s component increases as Cu content rises from $x=0.15$ to $x=0.25$.\\
D. The overall electrical resistivity of the ceramics is primarily governed by the grain resistance rather than the grain-boundary resistance.\\
E. Between the $x=0.15$ and $x=0.25$ compositions, the grain-boundary activation energy decreases while the grain activation energy decreases.\\
F. The average grain size exceeds 12\,\textmu m.

\smallskip
\textbf{Answer: A, B, C.} The withheld grain size is 10.22\,\textmu m (B holds, F fails). The study attributes the tetragonal phase to Jahn--Teller distortion of Cu$^{2+}$ at B sites (A), reports an increase in the oxygen-vacancy component from 35.8\% to 40.1\% (C), identifies grain-boundary resistance as dominant (D fails), and reports $E_a(R_{gb})$ falling from 0.253 to 0.229\,eV while $E_a(R_g)$ rises from 0.113 to 0.195\,eV (E fails).\\
\textbf{GPT-6-Astra selected A, B, C, E} (score 0); \textbf{GPT-6-Sol selected all six options} (score 0).
\end{examplebox}

\end{document}